\documentclass{article}

     \usepackage[final]{neurips_2024}

\usepackage[utf8]{inputenc} % allow utf-8 input
\usepackage[T1]{fontenc}    % use 8-bit T1 fonts
\usepackage{hyperref}       % hyperlinks
\usepackage{url}            % simple URL typesetting
\usepackage{booktabs}       % professional-quality tables
\usepackage{amsfonts}       % blackboard math symbols
\usepackage{nicefrac}       % compact symbols for 1/2, etc.
\usepackage{microtype}      % microtypography
\usepackage{amssymb}
\usepackage{amsmath}
\usepackage{amsthm}
\usepackage{cleveref}
\usepackage{autonum}
\usepackage{MnSymbol} 

\usepackage{caption}
\usepackage{url}
\usepackage{lettrine}
\usepackage{framed}
\usepackage{color}

\usepackage{algorithm,algpseudocode}

\algnewcommand{\algorithmicforeach}{\textbf{for each}}
\algdef{SE}[FOR]{ForEach}{EndForEach}[1]
  {\algorithmicforeach\ #1\ \algorithmicdo}% \ForEach{#1}
  {\algorithmicend\ \algorithmicforeach}% \EndForEach

\usepackage{overpic}
\usepackage{dsfont}
\usepackage{booktabs} 
\usepackage{bm}
\usepackage{bbm}
\usepackage{ifthen}
\usepackage{graphicx}
\usepackage{subcaption}
\usepackage{dsfont}
\usepackage{framed}
\usepackage{overpic}
\usepackage[svgnames]{xcolor}
\newtheorem{theorem}{Theorem}
\usepackage{apxproof}
\newtheorem{proposition}{Proposition}
\newtheorem{definition}{Definition}

\newtheorem{assumption}{Assumption}
\newtheorem{corollary}{Corollary}[theorem]
\hypersetup{
	colorlinks=true,
	linkcolor=blue,
	urlcolor=red,
	citecolor=DarkBlue,
	linkbordercolor={0 0 1}
}

\newtheorem{observation}{Observation}
\newtheoremrep{theorem}{Theorem}
\newcommand{\ocs}{\textsc{OCS}}

\renewcommand{\algorithmicrequire}{\textbf{Input:}}
\renewcommand{\algorithmicensure}{\textbf{Output:}}

\newcommand{\algorithmicinit}{\textbf{Initialize:}}
\newcommand{\INIT}{\item[\algorithmicinit]}

\newcommand{\edit}[1]{{}}
\theoremstyle{definition}

\title{Optimal Algorithms for Online Convex Optimization with Adversarial Constraints}

\author{%
  Abhishek Sinha, Rahul Vaze \\
 School of Technology and Computer Science \\
  Tata Institute of Fundamental Research \\
  Mumbai 400005, India \\
  \texttt{abhishek.sinha@tifr.res.in}, 
  \texttt{rahul.vaze@gmail.com}
}

\begin{document}

\maketitle

%\section{Abstract}
\begin{abstract}
%Constrained online convex optimization problem (COCO) is considered where after an online policy chooses an action $x_t$ on round $t$, the adversary reveals a convex cost function $f_t$, and a set of $k$ convex constraints of the form $g_{t,i}(x) \leq 0, i \in [k]$. The cost function $f_t$ and the constraint functions $g_{t,i}$'s could change arbitrarily with time, and no information about the future functions is assumed to be available. 
A well-studied generalization of the standard online convex optimization (OCO) framework is constrained online convex optimization (COCO). In COCO, on every round, a convex cost function and a convex constraint function are revealed to the learner after it chooses the action for that round. The objective is to design an online learning policy that simultaneously achieves a small regret while ensuring a small cumulative constraint violation (CCV) against an adaptive adversary interacting over a horizon of length $T$. A long-standing open question in COCO is whether an online policy can simultaneously achieve $O(\sqrt{T})$ regret and $\tilde{O}(\sqrt{T})$ CCV without any restrictive assumptions. For the first time, we answer this in the affirmative and show that a simple first-order policy can simultaneously achieve these bounds. Furthermore, in the case of strongly convex cost and convex constraint functions, the regret guarantee can be improved to $O(\log T)$ while keeping the CCV bound the same as above.
We establish these results by effectively combining adaptive OCO policies as a blackbox with Lyapunov optimization - a classic tool from control theory. Surprisingly, the analysis is short and elegant. 
%A potential function-based proof technique is presented that reveals a connection between regret and certain sequential inequalities through a novel decomposition result. The resulting guarantees either match the best-known results or provide new results in certain cases.
%We show that optimal performance bounds can be achieved by solving the surrogate problem using \emph{any} adaptive OCO policy that enjoys a standard data-dependent regret bound.  We conclude the paper by highlighting the application of the above framework to online multi-task learning and network control problems.
\end{abstract}
\section{Introduction} \label{intro}
Online convex optimization (OCO) is a standard framework for modelling and analyzing a broad family of online decision problems under uncertainty. In the OCO problem, on every round $t$, an online policy first selects an action $x_t$ from a closed and convex admissible set (\emph{a.k.a.} decision set) $\mathcal{X}.$ Then the adversary reveals a convex cost function $f_t$, resulting in a cost of $f_t(x_t)$. 
The goal of an online policy is to choose an admissible action sequence $\{x_t\}_{t=1}^T$ so that its cumulative cost is not much larger than that of any fixed admissible action chosen in hindsight. In particular, the objective is to minimize the static regret defined below
%\vspace{-0.1in}
\begin{eqnarray} \label{intro-regret-def}
	\textrm{Regret}_T \equiv \sup_{\{f_t\}_{t=1}^T} \sup_{x^\star \in \mathcal{X}} \textrm{Regret}_T(x^\star), ~\textrm{where~}\textrm{Regret}_T(x^\star) \equiv \sum_{t=1}^T f_t(x_t) - \sum_{t=1}^T f_t(x^\star).
\end{eqnarray}
 The term \emph{static} refers to using a fixed benchmark, specifically only one action $x^\star$ throughout the horizon of length $T$. 
 
 In this paper, we consider a generalization of the standard OCO framework. In this problem, on every round $t,$ the online policy first chooses an admissible action $x_t \in \mathcal{X},$ 
 and then the adversary chooses a convex cost function $f_t: \mathcal{X} \to \mathbb{R}$ and $k$ constraints of the form $g_{t,i}(x) \leq 0, \ i \in [k],$ where $g_{t,i}: \mathcal{X} \to \mathbb{R}$ is a convex function for each $i \in [k]$\footnote{Notations: For any natural number $n$, we define $[n] \equiv \{1,2,\ldots, n\}.$ For any real number $z$, we define $(z)^+ \equiv \max(0,z).$}. Since $g_{t, i}$'s are revealed after the action $x_t$ is chosen, an online policy need not necessarily take feasible actions on each round, and the obvious metric of interest in addition to \eqref{intro-regret-def} is the total cumulative constraint violation (CCV) $\mathbb{V}(T)$ defined as 
 %\vspace{-0.1in}
  \begin{eqnarray} \label{intro-gen-oco-goal}
 	\textrm{CCV}_T \equiv \mathbb{V}(T) = \max_{i=1}^k \mathbb{V}_i(T)\quad  \text{where} \quad \mathbb{V}_i(T)  = \sum_{t=1}^T (g_{t,i}(x_t))^+. 
	\end{eqnarray}
Let $\mathcal{X}^\star$ be the feasible set consisting of all admissible actions that satisfy all constraints $g_{t,i}(x) \leq 0, \ i \in [k], t\in [T]$. Under the standard assumption that $\mathcal{X}^\star$ is not empty, the goal is to design an online policy to simultaneously achieve a small regret \eqref{intro-regret-def} with respect to any admissible benchmark $x^\star \in \mathcal{X}^\star$ and a small CCV \eqref{intro-gen-oco-goal}. We refer to this problem as the constrained OCO (COCO). The assumption $\mathcal{X}^\star \neq \emptyset $ will be relaxed in Section \ref{simul_constr} for the  Online Constraint Satisfaction (OCS) problem where the cost functions are set to zero, and the objective is to minimize just the CCV.
%Compared to the stringent condition that 
%$\mathcal{X}^\star \neq \emptyset$, a relaxed condition is called the $S$-feasible benchmark, where the fixed action $x^\star$ in an $S$-feasible benchmark enjoys the property that the sum of the constraint functions evaluated at $x^\star$ over any consecutive sequence of $S \geq 1$ rounds is non-positive. Clearly, $S=1$ corresponds to  $\mathcal{X}^\star \neq \emptyset$. We consider both $S=1$ and general $S$-feasible benchmarks in this paper.

%control both the regret and the cumulative violation penalty optimally. This problem departs from the celebrated OCO framework because of the instantaneous constraints. This begets a natural question - Is it possible to efficiently reduce the constrained problem to a standard OCO problem while obtaining the optimal regret and cumulative violation bounds? In this paper, we answer this question affirmatively in a constructive fashion. 
%In this paper, our objective is to design a class of efficient online learning policies to simultaneously minimize the regret and the cumulative constraint violations.
%In addition to having a small regret, we require the cumulative constraint violations to be sublinear in time. 

%We consider the problem of Online Convex Optimization with time-varying instantaneous online constraints where the cost and constraint functions could be chosen adversarially. 
%\paragraph{Applications:} 
COCO arises in many applications, including online portfolio optimization with risk constraints, resource allocation in cloud computing with time-varying demands, pay-per-click online ad markets with budget constraints \citep{georgios-cautious}, online recommendation systems, dynamic pricing, revenue management, robotics and path planning problems, and multi-armed bandits with fairness constraints \citep{sinha2023banditq}. 
%Control policies for emerging applications such as autonomous vehicles face multi-dimensional constraints at each time, resulting from the requirement to stay on the course, speed constraints, collision avoidance constraints, and traffic regulatory constraints \citep{feng2023dense}. Offline optimization problems with a huge set of constraints can also be conveniently formulated in this framework. 
The necessity for revealing the constraints sequentially may also arise, \emph{e.g.,} in communication-limited settings, where it might be infeasible to reveal all constraints defining the feasible set at a time (\emph{e.g.,} combinatorial auctions). See Section \ref{expts} for an application of the COCO framework in fraud detection which involves binary classification with a highly-imbalanced dataset.  
\vspace{-0.1in}
\subsection{Related Work} \label{related}
%The general problem has remained open for a long time. It was conjectured that designing a policy with both sublinear regret and sublinear constraint violations is impossible without additional assumptions \cite{neely2017online}. 
%In fact, the authors in  \cite{mannor2009online} even commented that ``\ldots it is unlikely that such a reduction is possible.''
\paragraph{Unconstrained OCO:}
In a seminal paper, \citet{zinkevich2003online} showed that for solving \eqref{intro-regret-def}, the ubiquitous projected online gradient descent (OGD) policy achieves an $O(\sqrt{T})$ regret for convex cost functions with uniformly bounded sub-gradients. A number of follow-up papers proposed adaptive and parameter-free versions of OGD 
%that do not need to know any non-causal information 
\citep{hazan2007adaptive, orabona2018scale}. See \citet{orabona2019modern, hazan2022introduction} for textbook treatments of the OCO framework and associated algorithms.

%However, these lines of work do not consider additional constraints - a problem which has been systematically explored only recently (see Table \ref{gen-oco-review-table} for a brief summary). The constraint functions could either be known \emph{a priori} or revealed sequentially along with the cost functions. 
%\paragraph{Constrained OCO (COCO):} 
 {\bf Constrained OCO (COCO): (A) Time-invariant constraints:} A number of papers considered COCO with time-invariant constraints, \emph{i.e.,} $g_{t,i} = g_i, \forall \ t$ \citep{yuan2018online, jenatton2016adaptive, mahdavi2012trading, yi2021regret}.  These works assume that the functions $g_i$'s are known to the policy \emph{a priori}. However, they allowed the policy to remain infeasible on any round to avoid the costly projection step of the vanilla projected OGD  policy. Their main objective was to design an \emph{efficient} policy (avoiding the explicit projection step) with a small regret and CCV. 

{\bf (B) Time-varying constraints:} Solving the COCO problem when the constraint functions, \emph{i.e.}, $g_{t,i}$'s, change arbitrarily with time $t$ is more challenging. In this case,  except for 
 \cite{neely2017online} and \cite{georgios-cautious}, most of the prior works 
 %assumes that the policy has access to some non-causal information, \emph{e.g.,} a uniform upper bound to the norm of the future gradients of $f_t,g_{t,i}$. This non-causal information is used by their proposed policies to 
 construct some Lagrangian function and then update the primal and dual variables \citep{yu2017online, pmlr-v70-sun17a, yi2023distributed}. However, the performance bounds obtained with this approach remain suboptimal.
   Both \citet{neely2017online} and \cite{georgios-cautious} use the drift-plus-penalty (DPP) framework introduced by \citet{neely2010stochastic} to solve the constrained problem under various assumptions. In particular,
  \citet{neely2017online} proposed a DPP-based policy for COCO  upon assuming the Slater's condition, \emph{i.e.,} $g_{t,i}(x^\star) < -\eta$, for some $\eta>0$ $\forall i,t$. Clearly, this condition precludes the important case of non-negative constraint functions (\emph{e.g.,} constraint functions of the form $\max(0, g_t(x))$). Furthermore, the bounds obtained upon assuming Slater's condition depend inversely with the Slater's constant $\eta$ (usually hidden under the big-Oh notation). Since $\eta$ could be arbitrarily small, these bounds could be arbitrarily loose. 
  %Furthermore, a sublinear violation bound obtained upon assuming Slater's condition is loose by a quantity that increases \emph{linearly} with the horizon-length $T$ compared to a sublinear violation bound obtained without this assumption.    
%(which arises, e.g., upon a $\max(0,\cdot)$ operation with a given constraint). 
%Moreover, the regret bound presented in \cite{neely2017online} diverges to infinity as $\eta \searrow 0.$ 
 \cite{georgios-cautious} extended \cite{neely2017online}'s result by considering a weaker form of the feasibility assumption without assuming Slater's condition. 
% they show that a DPP-based policy achieves a regret $\mathcal{R}_T = O(ST/V + \sqrt{T})$ and CCV ${\mathbb V}(T) = O(\sqrt{VT})$. Here, $V$ is an adjustable parameter that can take any value in $[S, T).$ Hence, \emph{a priori}, their algorithm needs to know the value of the parameter $S,$ which, unfortunately, depends on the online constraints.
%It can be seen that the violation penalty bound achieved by their policy is at least $O(T^{3/4}),$ which is suboptimal. 
Furthermore, although these DPP-based results are interesting, they have not been able to provide improved regret or CVV bounds 
when the cost functions $f_t$'s are strongly convex because of the linearization step inherent in this approach.

In a recent paper, \citet{guo2022online} considered COCO and obtained the best-known prior results without assuming Slater's condition. However, in addition to yielding sub-optimal bounds, their policy is quite computationally intensive since it requires solving a convex optimization problem on each round. Compared to this, all policies proposed in this paper take only a single gradient-descent step and perform only one Euclidean projection on each round. 
%Moreover, it is unclear how to extend \citet{guo2022online}'s policy to the more general $S$-feasible benchmark, where it is necessary to compensate for constraint violations at some rounds with strictly satisfying constraints on some other rounds. 
Please refer to Table \ref{gen-oco-review-table} for a brief summary of the results and Section \ref{app:comparisonpolicies} in the Appendix for a qualitative comparison.
%inefficient as, instead of performing a single gradient-descent step per round (as in our and most of the previous algorithms), their algorithm needs to solve a general convex optimization problem at every round. Moreover, their algorithm needs access to the full description of the constraint function $g_t(\cdot)$ for the optimization step, whereas ours and most of the previous algorithms need to know only the gradient and the value of the constraint function for the current action $x_t$. 
%In a recent paper, \cite{yi2023distributed} consider the same problem in a distributed setup and derive tighter bounds upon assuming Slater's condition.
The COCO problem has been considered in the {\it dynamic} setting as well  \citep{chen2018bandit, cao2018online, vazecocowiopt2022, liu2022simultaneously} where the benchmark $x^\star$ in \eqref{intro-regret-def} is replaced by $x_t^\star$ that is also allowed to change its actions over time. However, we focus our attention on achieving the optimal performance bounds for the static version.
%\paragraph{The Online Constraint Satisfaction (\textsc{OCS}) Problem:} %In addition to studying the above problem, we also introduce a new but related problem, 
A special case of COCO is the 
\textsc{Online Constraint Satisfaction} (\ocs) problem that does not involve any cost function, \emph{i.e.,} $f_t=0, \ \forall t,$ and the only object of interest is the CCV. The \ocs ~problem becomes especially interesting in the setting where the feasible set may be empty.
  \begin{table*}[t]
%\hspace{-30pt}
  \title{Summary of the results for the constrained OCO problem}
  %\centering
  \hspace{-40pt}
  \begin{tabular}{llllll}
    \toprule
    %\multicolumn{2}{c}{Part}                   \\
   % \cmidrule(r){1-2}
   \small { Reference}  & \small {Regret} & \small {CCV} & \small {Complexity per round}& \small {Assumptions} \\
    \midrule
  %  a & b& c& d & e  \\
      \small {\citet{mahdavi2012trading}}  & \small {$O(\sqrt{T})$} & \small {$O(T^{\nicefrac{3}{4}})$} & \small {Projection} & \small{Time-invariant constraints} \\
    \small {\citet{jenatton2016adaptive}}  & \small {$O(T^{\max(\beta, 1-\beta)})$} & \small {$O(T^{1-\beta/2})$} & \small {Projection} & \small{Time-invariant constraints} \\
    \small {\citet{pmlr-v70-sun17a}}  & \small {$O(\sqrt{T})$} & \small {$O(T^{\nicefrac{3}{4}})$}& \small {Bregman Projection} & \small -  \\
    \small {\citet{neely2017online}}  & \small {$O(\sqrt{T})$} & \small {$O(\sqrt{T})$} & \small {Conv-OPT} & \small {Slater condition} \\
    %    \small {\citet{yu2017online}}  & \small {$O(\sqrt{T})$} & \small {$O(\sqrt{T})$} & \small {Conv-OPT} & \small {Slater condition} \\
  \small {\citet{yuan2018online}} & \small {$O(T^{\max(\beta, 1-\beta)})$} & \small {$O(T^{1-\beta/2})$ }  & \small {Projection} & \small{Time-invariant constraints} \\
      \small {\citet{yu2020low}}  & \small {$O(\sqrt{T})$} & \small {$O(1)$} & \small {Conv-OPT} & \small {Slater \& Time-invariant constraints} \\
 % \citet{yu2017online} & Stochastic & $O(\sqrt{T})$ & $O(\sqrt{T})$& OGD+drift+penalty & Slater condition \\
  \small {\citet{yi2021regret}} & \small {$O(T^{\max(\beta, 1-\beta)})$} & \small {$O(T^{(1-\beta)/2})$} & \small {Conv-OPT} & \small{Time-invariant constraints} \\ 
  \small {\citet{yi2022regret}}  & \small {$O(T^{\beta})$} & \small {$O(T^{1-\beta/2})$} & \small {Projection} & \small {Strongly convex cost} \\
    \small {\citet{guo2022online}}  & \small {$O(\sqrt{T})$} & \small {$O(T^{\nicefrac{3}{4}})$} & \small {Conv-OPT} & - \\
        \small {\citet{guo2022online}}  & \small {$O(\log T)$} & \small {$O(\sqrt{T \log T})$} & \small {Conv-OPT} & \small {Strongly convex cost} \\ 
  \small {\citet{yi2023distributed}}  & \small {$O(T^{\max(\beta, 1-\beta)})$} & \small {$O(T^{1-\beta/2})$} & \small {Conv-OPT} & - \\
    \small {\citet{yi2023distributed}} & \small {$O(\log(T))$} & \small {$O(\sqrt{T \log T})$} & \small {Conv-OPT} & \small {Strongly convex cost} \\
    %\citet{georgios-cautious} & Adversarial, convex & $S$ & $O(\sqrt{ST})$ & OGD & Known $S$ \\
               %\citet{guo2022online} & Adversarial, strongly-convex & $1$ & $O(\sqrt{T \log T})$ & Convex opt. each round & -do-, Known $\alpha$ \\
   %      \small {\citet{georgios-cautious}}  & &\ \textcolor{red}{($S$,$O(\sqrt{ST})$)}& \small {OGD} & \small {Known $S$} \\
  %\small {\textbf{This paper}}  & \small {$O(\sqrt{T})$} & \small {$O(T^{3/4})$} & \small {Ad-OCO} & - \\
     \small {\textbf{This paper}} & \small {$O(\sqrt{T})$} & \small {$O(\sqrt{T}\log T)$} & \small {Projection} & -\\
 \small {\textbf{This paper}} &   \small {$O(\log T)$} & \small {$O(\sqrt{T\log T})$} &\small {Projection} & \small{Strongly convex cost} \\
 \small {\textbf{This paper}} & \small {$O(\log T)$} & \small {$O(\frac{\log T}{\alpha})$} &\small {Projection} & \small {Strongly convex cost, $\textrm{Regret}_T \geq 0,$} \\
       \bottomrule
  \end{tabular}
  \vspace{5pt}
  \caption{\small{Summary of the results on COCO. Unless stated otherwise, we assume arbitrary time-varying convex constraints and convex cost functions. In the above table, $0\leq \beta \leq 1$ is an adjustable parameter, $\alpha$ is the strong convexity parameter of the strongly convex cost functions. Conv-OPT refers to solving a constrained convex optimization problem on each round. Projection refers to the Euclidean projection operation on the convex set $\mathcal{X}$. For typical convex sets (\emph{e.g.,} Euclidean box, probability simplex), projection operations are substantially more efficient than solving a constrained convex optimization problem.}}
    \label{gen-oco-review-table}
\end{table*}
\subsection{Our Contributions} \label{contribution}
In this paper, we consider both COCO and $\ocs$ problems and make the following contributions. 
\begin{enumerate}
%\vspace{-.1in}
	\item 
	We propose an efficient first-order policy that simultaneously achieves $O(\sqrt{T})$ regret and $O(\sqrt{T}\log T)$ CCV for the COCO problem. Our result breaks the long-standing $O(T^{\nicefrac{3}{4}})$ barrier for the CCV and matches the lower bound (derived in Theorem \ref{thm:lbcoco}, previously missing from the literature) up to a logarithmic term. For strongly convex cost functions, the regret guarantee is improved to $O(\log T)$ while  keeping the CCV bound the same as above. Under an additional assumption that the regret is non-negative, we obtain a further improved logarithmic CCV bound in the strongly convex setting (see Table \ref{gen-oco-review-table}). 
\item We additionally consider a special case of the COCO problem, called Online Constraint Satisfaction (OCS), under relaxed feasibility assumptions and obtain sub-linear CCV bounds. 

\item On the algorithmic side, our policy simply runs an adaptive first-order OCO algorithm as a blackbox on a specially constructed convex surrogate cost function sequence. On every round, the policy needs to compute only two gradients and an Euclidean projection. This is way more efficient compared to the policies proposed in the previous works \citep{guo2022online, neely2017online}, which need to solve expensive convex optimization problems on each round while yielding sub-optimal bounds. Furthermore, in the special case of time-invariant constraints, our results yield an efficient first-order OCO policy with competitive regret and CCV bounds \citep{mahdavi2012trading, jenatton2016adaptive, yi2021regret}. 

\item Our results are obtained by introducing a crisp and elegant potential function-based algorithmic technique for simultaneously controlling the regret and the CCV. In brief, the regret and CCV bounds are derived from a single inequality that arises from plugging in off-the-shelf adaptive regret bounds in a new regret decomposition result (Eqn.\ \eqref{gen-reg-decomp}). This new analytical technique might also be of independent interest. 
\item Finally, in Section \ref{expts}, we evaluate the practical performance of our algorithm in the online credit card fraud detection problem with a highly imbalanced dataset.
\end{enumerate}
\section{The Constrained OCO (COCO) Problem} \label{gen_oco}
\subsection{Assumptions}  \label{assump}
%In this section, we list the general assumptions which apply to both the \ocs ~problem and the COCO, described later in Section \ref{gen_oco}. Since the \ocs ~problem does not contain any cost function, the cost functions mentioned below necessarily refer to COCO only.
We now state the assumptions considered in this paper. These assumptions are standard in literature on the COCO problem \citep{guo2022online, yi2021regret, neely2017online}.
\begin{assumption}[Convexity] \label{cvx}
	The cost function $f_t: \mathcal{X} \mapsto \mathbb{R}$ and the constraint function $g_{t,i}: \mathcal{X} \mapsto \mathbb{R}$ are convex for all $t\geq 1, i\in [k]$. The admissible set (\emph{a.k.a.} the decision set or the action set) $\mathcal{X} \subseteq \mathbb{R}^d$ is closed and convex and has a finite Euclidean diameter $D$. 
 %Moreover, $D$ is known ahead of time.
\end{assumption}
%\vspace{-0.18in}
\begin{assumption}[Lipschitzness] \label{bddness}
 %We have $\textrm{diam}(\mathcal{X}) \leq D, ||\nabla f_t(x)||_2 \leq G/2, \textrm{and}~ ||\nabla g_t(x))||_2 \leq G/2,~\forall t, \forall x\in \mathcal{X}$ for some finite constants $D$ and $G.$ If the functions are not necessarily differentiable, we require that the maximum magnitude of the subgradients be bounded accordingly.  Each
All cost functions $\{f_t\}_{t\geq 1}$ and the constraint functions $\{g_{t,i}\}_{i\in [k], t\geq 1}$'s are $G$-Lipschitz. In other words, for any $x, y \in \mathcal{X},$ we have 
 \begin{eqnarray*}
 	|f_t(x)-f_t(y)| \leq G||x-y||,~
 	|g_{t,i}(x)-g_{t,i}(y)| \leq G||x-y||, ~\forall t\geq 1, i\in [k].
 \end{eqnarray*}
	\end{assumption}

	Unless specified otherwise, the norm $||\cdot||$ will refer to the standard Euclidean norm and $\nabla f$ will refer to an arbitrary subgradient of a convex function $f$. Assumption \ref{bddness} implies that the $\ell_2$-norm of the (sub)gradients of the cost and constraint functions are uniformly upper-bounded by $G$ over the admissible set $\mathcal{X}.$ Finally, we make the following feasibility assumption about the constraint functions.
\begin{assumption}[Feasibility] \label{feas-constr}
	There exists a feasible action $x^\star \in \mathcal{X} $ s.t. $g_{t,i}(x^\star) \leq 0, \forall t, i.$ The feasible set $\mathcal{X}^\star$ is defined to be the set of all feasible actions. The feasibility assumption implies that $\mathcal{X}^\star \neq \emptyset.$
\end{assumption}
The feasibility assumption distinguishes the cost functions from the constraint functions and is commonly assumed in the literature \citep{guo2022online, neely2017online, yu2016low,yuan2018online,yi2023distributed, georgios-cautious}.  In Section \ref{simul_constr}, we will consider a constraint-only variant of the problem where the feasibility assumption (Assumption \ref{feas-constr}) will be relaxed. 
See Appendix \ref{app:assumptions} for a brief discussion on the assumptions.

\paragraph{Remarks:} On each round, multiple constraints of the form $g_{t,i}(x) \leq 0, i\in [k]$ can be replaced by a single new constraint $g_t(x) \leq 0$
%Multiple constraints per round can be reduced to a single constraint by simply clipping each of the constraints and 
%by replacing them with a new constraint 
where the constraint function $g_t$ is defined to be the pointwise maximum of the given constraints, \emph{i.e.,} 
 $g_t(x) \equiv \max_{i=1}^k g_{t,i}(x), x \in \mathcal{X}.$ It is easy to verify that if each of the constraint functions $\{g_{t,i}\}_{i=1}^k$ satisfies the above assumptions, then the constraint function $g_t$ defined above also satisfies the assumptions. Hence,  throughout this section and without loss of generality, we will assume that only one constraint function is revealed on each round. That being said, under the relaxed feasibility assumption in Section \ref{simul_constr}, this trick does not work and there we will need to consider the full set of $k$ constraint functions.  
\vspace{-0.1in}
\subsection{Online Policy for COCO} \label{policy}
Recall that compared to the standard OCO problem where the only objective is to minimize the Regret \citep{hazan2022introduction}, in COCO, our objective is twofold: to \emph{simultaneously} control the Regret \emph{and} the CCV. See Section \ref{prelims} in the Appendix for preliminaries on the OCO problem and some standard results which will be useful in our analysis. In the following, we 
propose a Lyapunov function-based policy 
%recently proposed by \citet{sinha2023playing}. While they were able to prove an $O(\sqrt{T})$ regret and $O(T^{\nicefrac{3}{4}})$ CCV with a quadratic Lyapunov function, in this paper, we show that this technique can be generalized with a power-law Lyapunov function to 
that yields the optimal Regret and CCV bounds for the COCO problem. Although for simplicity, we assume that the horizon length $T$ is known, we can use the standard doubling trick for an unknown $T.$

\subsection{Design and Analysis of the Algorithm}
 
To simplify the analysis, we pre-process the cost and constraint functions on each round as follows.

\vspace{5pt}

%the queue-lengths evolve as in Eqn.\ \eqref{q-ev}, where 
\hrule
\textbf{Pre-processing:}
On every round, we first clip the negative part of the constraint function to zero by passing it through the standard ReLU unit. Then, we scale both the cost and constraint functions by a positive factor $\beta,$ which will be determined later. In other words, 
 we work with the pre-processed inputs $\tilde{f}_t \gets \beta f_t, \tilde{g}_t \gets \beta (g_t)^+.$ Hence, the pre-processed functions are $\beta G$-Lipschitz and $\tilde{g}_t \geq 0, \forall t.$  
 \vspace{5pt}

\hrule 
In the following, we derive the Regret and CCV bounds for the pre-processed functions. The bounds for the original problem are obtained upon scaling the results back by $\beta^{-1}$ in the final step.
\subsubsection{Defining the Surrogate Cost Functions} 
%we can obtain the optimal regret and violation bound for the COCO problem with convex cost and convex constraint functions.

Let $Q(t)$ denote the CCV for the pre-processed constraints up to round $t.$ Clearly, $Q(t)$ satisfies the simple recursion $Q(t)=Q(t-1)+\tilde{g}_t(x_t), t\geq 1, $ with $Q(0)=0$. Recall that one of our objectives is to make $Q(t)$ small.
Towards this, let $\Phi: \mathbb{R}_+ \mapsto \mathbb{R}_+$ be any non-decreasing differentiable convex potential (Lyapunov) function such that $\Phi(0)=0.$ Using the convexity of $\Phi(\cdot),$ we have
%which generalizes Eqn.\ \eqref{dr-bd}:
%Also, for the sake of simplicity, we assume that the maximum magnitude of the constraint violation is upper bounded by $F=1$. 
%Since the function $ h: x \to x^n$ is convex, we have 
%\begin{eqnarray*}
%	\Phi(t)= Q^n(t) = \big(Q(t-1)+g_t(x_t)\big)^n \leq Q^{n}(t-1) + n Q^{n-1}(t) g_t(x_t). 
%	\end{eqnarray*}
%
\begin{eqnarray} \label{dr-bd-gen}
	\Phi(Q(t))  &\leq& \Phi(Q(t-1)) + \Phi'(Q(t))(Q(t)-Q(t-1)) \nonumber \\
&=& \Phi(Q(t-1)) + \Phi'(Q(t)) \tilde{g}_t(x_t). 
\end{eqnarray}
Hence, the change (\emph{drift}) of the potential function $\Phi(Q(t))$ on round $t$ can be upper bounded as 
\begin{eqnarray} \label{drift_ineq_new}
	\Phi(Q(t))-\Phi(Q(t-1)) \leq \Phi'(Q(t)) \tilde{g}_t(x_t). 
\end{eqnarray}
Recall that, in addition to controlling the CCV, we also want to minimize the cumulative cost $\sum_{t=1}^T f_t(x_t)$ (which is equivalent to the regret minimization). Inspired by the stochastic \emph{drift-plus-penalty} framework of \citet{neely2010stochastic}, we combine these two objectives to a single objective of minimizing a sequence of surrogate cost functions $\{\hat{f}_t\}_{t=1}^T$ which are obtained by taking a positive linear combination of the drift upper bound \eqref{drift_ineq_new} and the cost function. More precisely, we define
%Inspired by the stochastic \emph{drift-plus-penalty} framework of \citet{neely2010stochastic}, on round $t$, we attempt to minimize a surrogate cost function $\hat{f}_t : \mathcal{X} \to \mathbb{R},$ obtained by adding the scaled cost function $f_t$ to a functional form of the drift upper bound \eqref{drift_ineq_new} as defined below:
\begin{eqnarray} \label{surrogate_new}
	\hat{f}_t(x):= V\tilde{f}_t(x)+ \Phi'(Q(t)) \tilde{g}_t(x), ~~ t \geq 1. 
\end{eqnarray}
In the above, $V$ is a suitably chosen non-negative parameter to be determined later. In brief, the proposed policy for COCO, described in Algorithm \ref{coco_alg}, simply runs an adaptive OCO policy on the surrogate cost function sequence $\{\hat{f}_t\}_{t\geq 1}$, with a specific choice of the potential function $\Phi(\cdot)$, the parameter $V$, and step-size sequence $\{\eta_t\}_{t\geq 1}$, as dictated by the following analysis.
\begin{algorithm}[tb]
   \caption{Online Policy for COCO}
   \label{coco_alg}
\begin{algorithmic}[1]
   \State {\bfseries Input:} Sequence of convex cost functions $\{f_t\}_{t=1}^T$ and constraint functions $\{g_t\}_{t=1}^T,$ $G=$ a common Lipschitz constant, $T=$ Horizon length,
   %an upper bound $G$ to the Euclidean norm of their (sub)gradients, 
    $D=$ Euclidean diameter of the admissible set $\mathcal{X},$ $\mathcal{P}_\mathcal{X}(\cdot)=$ Euclidean projection operator on the set $\mathcal{X}$ 
     \State {\bfseries Parameter settings:} 
     \begin{enumerate}
     	\item \textbf{Convex cost functions:} $\beta = (2GD)^{-1}, V=1, \lambda = \frac{1}{2\sqrt{T}}, \Phi(x)= \exp(\lambda x)-1.$
     
    \item \textbf{$\alpha$-strongly convex cost functions:} $\beta =1, V=\frac{8G^2 \ln(Te)}{\alpha}, \Phi(x)= x^2.$
    \end{enumerate}
     %$ \alpha=\frac{1}{2GD}, n=\max(2, \lceil \ln T \rceil), V=(n-1)^{n-1}T^{\frac{n-1}{2}}, \Phi(x)=x^n.$ 
%   \REPEAT
  \State {\bfseries Initialization:} Set $ x_1 \in \mathcal{X}$ arbitrarily, $Q(0)=0$.
   \ForEach{$t=1:T$}
   \State Play $x_t,$ observe $f_t, g_t,$ incur a cost of $f_t(x_t)$ and constraint violation of $(g_t(x_t))^+$
   \State $\tilde{f}_t \gets \beta f_t, \tilde{g}_t \gets \beta \max(0,g_t).$
   \State $Q(t)=Q(t-1)+\tilde{g}_t(x_t).$
   \State Compute (sub)gradient $\nabla_t = \nabla \hat{f}_t(x_t),$ where the surrogate function $\hat{f}_t$ is defined in Eqn.\ \eqref{surrogate_new}
   \State $x_{t+1} = \mathcal{P}_\mathcal{X}(x_t - \eta_t \nabla_t)$, where 
   \begin{eqnarray*}
   \eta_t =\begin{cases}
   	\frac{\sqrt{2}D}{2\sqrt{\sum_{\tau=1}^{t} ||\nabla_\tau||_2^2}}, ~&~\textrm{for convex costs (AdaGrad stepsizes)} \\
   	\frac{1}{\sum_{s=1}^t H_s}, ~ &~ \textrm{for strongly convex costs } (H_s \textrm{= strong convexity parameter of } f_s, s\geq 1) 
   	\end{cases}
   	\end{eqnarray*}
   	
%   \IF{$x_i > x_{i+1}$}
%   \STATE Swap $x_i$ and $x_{i+1}$
%   \STATE $noChange = false$
%   \ENDIF
   \EndForEach
%   \UNTIL{$noChange$ is $true$}
\end{algorithmic}
\end{algorithm}
%\begin{framed}
%\paragraph{Policy for COCO:} Run the AdaGrad algorithm \ref{ogd-policy} on the sequence of surrogate cost functions $\{\hat{f}_t\}_{t\geq 1}$ given by Eqn.\ \eqref{surrogate_new}.
%\end{framed}

\subsubsection{The Regret Decomposition Inequality}
Let $x^\star \in \mathcal{X}^\star$ be any feasible action guaranteed by Assumption \eqref{feas-constr}. Plugging in the definition of surrogate costs \eqref{surrogate_new} into the drift inequality \eqref{drift_ineq_new}, and using the fact that $g_\tau(x^\star)\leq 0, \forall \tau \geq 1,$ we have
%Working similarly as before, we have the following inequality
\begin{eqnarray*}
	\Phi(Q(\tau))-\Phi(Q(\tau-1)) + V(\tilde{f}_\tau(x_\tau)-\tilde{f}_\tau(x^\star)) 
	\leq \hat{f}_\tau(x_\tau) - \hat{f}_\tau(x^\star), ~ \forall \tau \geq 1.
\end{eqnarray*}
Summing the above inequalities for rounds $1\leq \tau \leq t$, and using the fact that $\Phi(0)=0,$ we obtain  
\begin{eqnarray} \label{gen-reg-decomp}
	\Phi(Q(t)) +V \textrm{Regret}_t(x^\star) \leq \textrm{Regret}_t'(x^\star), ~ \forall x^\star \in \mathcal{X}^\star,
\end{eqnarray}
where $\textrm{Regret}_t$ on the LHS and $\textrm{Regret}'_t$ on the RHS of \eqref{gen-reg-decomp} refer to the regret for learning the pre-processed cost functions $\{\tilde{f}_t\}_{t\geq 1}$ and the surrogate cost functions $\{\hat{f}_t\}_{t \geq 1}$ respectively. 
%First, 
We will use the following upper bound on the $\ell_2$-norm of the (sub)gradients $G_t$ of the surrogate cost function $\hat{f}_t$ defined in Eqn.\ \eqref{surrogate_new}:
\begin{eqnarray} \label{grad_bd_new}
	%||\nabla \hat{f}_t(x_t)||_2
	G_t \equiv ||\nabla \hat{f}_t(x_t)||
	\stackrel{(a)}{\leq} V||\nabla \tilde{f}_t(x_t)||+ \Phi'(Q(t))||\nabla \tilde{g}_t(x_t)|| 
	\stackrel{(b)}{\leq} \beta G\big(V+\Phi'(Q(t)\big),
\end{eqnarray}
where in $(a)$, we have used the triangle inequality for $\ell_2$ norms and in $(b)$, we have used the fact that all pre-processed functions are $\beta G$-Lipschitz. 
\subsubsection{Convex Cost and Convex Constraint Functions}
We now apply the regret decomposition inequality \eqref{gen-reg-decomp} to the case of convex cost and convex constraint functions.  
%without making any additional assumption beyond Assumption \ref{bddness} and Assumption \ref{slater}. 
%Since Algorithm \ref{coco_alg} uses the AdaGrad algorithm for learning the surrogate cost functions, from \eqref{cvx-reg-bd}, we need to upper bound the gradients of the surrogate functions to derive the regret bound. Towards this, 
Let us choose the regret-minimizing OCO subroutine for the surrogate cost functions to be the OGD policy with adaptive step sizes (a.k.a. \emph{AdaGrad}) described in part 1 of Theorem \ref{data-dep-regret} in the Appendix (see Algorithm \ref{coco_alg}). 
Plugging in the adaptive regret bound \eqref{cvx-reg-bd} on the RHS of \eqref{gen-reg-decomp}, setting $\beta=(2GD)^{-1},$ and using Eqn.\ \eqref{grad_bd_new}, we arrive at the following inequality valid for any $t \geq 1: $
\begin{eqnarray} \label{gen-fn-ineq}
		\Phi(Q(t)) +V \textrm{Regret}_t(x^\star) \leq \sqrt{\sum_{\tau=1}^t \big(\Phi'(Q(\tau))\big)^2} + V\sqrt{t}.
\end{eqnarray}
%The above inequality is obtained by first upper-bounding 
In deriving the above result, we have utilized simple algebraic inequalities $(x+y)^2 \leq 2(x^2+y^2)$ and $\sqrt{a+b} \leq \sqrt{a} + \sqrt{b}, a, b\geq 0.$ Now recall that the sequence $\{Q(t)\}_{t\geq 1}$ is non-negative and non-decreasing as $\tilde{g}_t\geq 0.$ Furthermore, the derivative $\Phi'(\cdot)$ is non-decreasing as the function $\Phi(\cdot)$ is assumed to be convex. Hence, bounding all terms in the summation on the RHS of \eqref{gen-fn-ineq} from above by the last term, we arrive at the following inequality for any feasible $x^\star \in \mathcal{X}^\star:$ 
\begin{eqnarray} \label{gen-fn-ineq2}
	\Phi(Q(t)) +V \textrm{Regret}_t(x^\star) \leq \Phi'\big(Q(t)\big)\sqrt{t} + V\sqrt{t}.
\end{eqnarray}
The simplified regret decomposition inequality \eqref{gen-fn-ineq2} constitutes the key step for the subsequent analysis.
%Where, to ease typing, we have assumed that the functions are scaled such that $2GD \leq 1$.
\paragraph{$\blacksquare$ Performance Analysis}
%~Analysis of Algorithm \ref{coco_alg}}
\paragraph{An exponential Lyapunov function:} We now derive the Regret and CCV bounds for the proposed  policy (Algorithm \ref{coco_alg}) by choosing $\Phi(\cdot)$ to be the exponential Lyapunov function: $\Phi(x)\equiv \exp(\lambda x)-1,$ where the parameter $\lambda \geq 0$ will be fixed later. 
%An analysis with a power-law potential function is given in Appendix \ref{power-law}, which also yields similar bounds. 
Clearly, the function $\Phi(\cdot)$ satisfies the required conditions for a Lyapunov function - it is a non-decreasing and convex function with $\Phi(0)=0.$ 
\paragraph{Bounding the Regret:}
%Since the  sequence $\{Q(t)\}_{t\geq 1}$ is non-negative and non-decreasing as the pre-processed constraints are non-negative, upper-bounding all terms in the summation of the RHS of \eqref{gen-fn-ineq} by the last term, we have the following regret bound for 
With the above choice for the Lyapunov function $\Phi(\cdot)$, Eqn.\ \eqref{gen-fn-ineq2} implies that for any feasible $x^\star \in \mathcal{X}^\star$ and for any $t \in [T],$ we have
\begin{eqnarray*}
	\exp(\lambda Q(t)) -1 + V \textrm{Regret}_t(x^\star) \leq \lambda \exp(\lambda Q(t)) \sqrt{t} + V\sqrt{t}.
\end{eqnarray*}
Transposing the first term on the above inequality to the RHS and dividing throughout by $V$, we have: 
\begin{eqnarray} \label{reg-bd-exp}
	\textrm{Regret}_t(x^\star ) \leq \sqrt{t}+\frac{1}{V}+ \frac{\exp(\lambda Q(t))}{V}(\lambda \sqrt{t}-1).
\end{eqnarray}
Choosing any $\lambda \leq  \frac{1}{\sqrt{T}},$ the last term in the above inequality becomes non-positive for any $t \in [T].$ Hence, for any $x^\star \in \mathcal{X}^\star$, we have the following regret bound
\begin{eqnarray} \label{regret-bd1}
	\textrm{Regret}_t(x^\star ) \leq \sqrt{t}+\frac{1}{V}. ~~ \forall t \in [T].
\end{eqnarray}
\paragraph{Bounding the CCV:} 
Since all pre-processed cost functions are $\beta G=(2D)^{-1}$-Lipschitz,
%with Lipschitz constant $\alpha G= (2D)^{-1}$, 
we trivially have $\textrm{Regret}_t(x^\star) = \sum_{s=1}^t (\tilde{f}_s(x_s) - \tilde{f}_s(x^\star)) \geq -\frac{Dt}{2D} \geq -\frac{t}{2}.$ Hence, from Eqn.\ \eqref{reg-bd-exp}, we have that for any $\lambda < \frac{1}{\sqrt{T}}$ and any $t \in [T]:$
\begin{eqnarray} \label{q-len-exp-bd}
	\frac{\exp(\lambda Q(t))}{V}(1-\lambda \sqrt{t}) \leq  2t + \frac{1}{V} 
	 \implies  Q(t) \leq  \frac{1}{\lambda}\ln\frac{1+2Vt}{1-\lambda \sqrt{t}}.
\end{eqnarray}
Choosing $\lambda=\frac{1}{2\sqrt{T}}, V=1,$ and scaling the bounds back by $\beta^{-1}\equiv 2GD,$ we arrive at our main result.
\begin{theorem} \label{main_result}
For the COCO problem with adversarially chosen $G$-Lipschitz cost and constraint functions, Algorithm \ref{coco_alg}, with $\beta=(2GD)^{-1}, V=1, \Phi(x)= \exp(\frac{x}{2\sqrt{T}})-1,$ yields the following Regret and CCV bounds for any horizon length $T \geq 1:$
%for any $T \geq 1$
\begin{eqnarray*}
 \textrm{Regret}_t \leq 2GD(\sqrt{t}+1), ~~\forall t \in [T], ~~\textrm{CCV}_T \leq 4GD\ln(2\big(1+2T)\big)\sqrt{T}.
 \end{eqnarray*}
 In the above, $D$ denotes the Euclidean diameter of the closed and convex admissible set $\mathcal{X}$.
\end{theorem}
 \subsubsection{Strongly Convex Cost and Convex Constraint Functions} \label{str-cvx-lin-cnst}
We now consider the setting where each of the cost functions $f_t, t\geq 1,$ is $\alpha$-strongly convex for some $\alpha>0$. The constraint functions $g_t$'s are assumed to be convex as before and not necessarily strongly convex. 
%For this case, however, we assume that the values of the parameters $\alpha$ and $G$ are known to the policy, which will be used for setting the parameter $V.$
%Note that the value of the strong-convexity parameter $\alpha$ needs not be known to the policy. 
%Also we make the standard assumption that $\textrm{diam}(\mathcal{X}) \leq D.$ Our objective is to derive better regret and constraint violation penalty bound for this case by making use of the $L^\star$ regret bound given in \cite[Theorem 4.25]{orabona2019modern}. Generalizing the policy described above, we now consider a sequence of surrogate penalty functions.
% \begin{eqnarray} \label{transformed_cost}
% 	 c_t'(x) = Vc_t(x) - 2Q(t)\langle a_t, x\rangle,  
% \end{eqnarray}
% for some constant $V>0$ that will be fixed later.
  %In Algorithm \ref{coco_alg}, 
  In this case, we choose the regret-minimizing OCO subroutine for the surrogate cost functions to be the OGD algorithm with the step-size sequence as given in part 2 of Theorem \ref{data-dep-regret} in the Appendix (see Algorithm \ref{coco_alg}).
  %, whose regret bound is given by Eqn.\  \eqref{str-cvx-reg-bd}.
  \iffalse
  following adaptive regret bound \cite[Theorem 4.1]{hazan2007adaptive}:
 \begin{eqnarray} \label{strong-cvx-regret}
 	\textrm{Regret}_T \leq \frac{1}{2}\sum_{t=1}^T \frac{G_t^2}{H_{1:t}},
 \end{eqnarray} 
 where $H_{1:t}$ is the sum of the strong convexity parameters of the first $t$ cost functions.
 \fi
% Now proceeding similarly as above, we have 
% \begin{eqnarray} \label{regret-bd}
% 	Q^2(t) + V \textrm{Regret}_t \leq \textrm{Regret}_t'(x^\star). 
% \end{eqnarray}
% Note that for the transformed sequence of functions, we have 
% \begin{eqnarray*}
% 	||\nabla c_t'(x_t)||^2 \leq 2V^2L^2 + 8Q(t)^2.
% \end{eqnarray*} 
 Since the cost functions are known to be $\alpha$-strongly convex, each of the surrogate cost functions \eqref{surrogate_new} is $V\alpha$-strongly convex. Hence, using the bound from Eqn.\ \eqref{grad_bd_new}, 
 %plugging in the upper bound of the norm of the gradient of the surrogate costs from  \eqref{grad_bd_new} into  \eqref{str-cvx-reg-bd}, 
 choosing the scaling parameter to be $\beta=1$, and simplifying the generic regret bound given by Eqn.\ \eqref{str-cvx-reg-bd}, we obtain the following regret bound for learning the surrogate cost functions $\{\hat{f}_s\}_{s\geq 1}$:
 %\vspace{-0.1in}
 \begin{eqnarray}\label{adaptive_str_cvx_bd}
 	\textrm{Regret}'_t(x^\star) \leq \frac{VG^2}{\alpha} (1+\ln(t)) + \frac{G^2}{\alpha V} \sum_{\tau=1}^t \frac{(\Phi'(Q(\tau)))^2}{\tau}, ~ x^\star \in \mathcal{X}. 
 \end{eqnarray}
 %\vspace{-0.05in}
In the above, we have used the standard bound for the Harmonic sum: $\sum_{\tau=1}^t \frac{1}{\tau} \leq 1+ \ln(t)$, as well as the fact that $(a+b)^2 \leq 2(a^2+b^2). $
 Substituting the bound \eqref{adaptive_str_cvx_bd} into the regret decomposition inequality \eqref{gen-reg-decomp}, and using the non-decreasing property of the sequence $\{Q(\tau)\}_{\tau \geq 1}$ and the derivative $\Phi'(\cdot)$, we obtain 
 %\vspace{-0.1in}
 \begin{eqnarray} \label{Gronwall-ineq}
 	\Phi(Q(t)) + V \textrm{Regret}_t(x^\star)  \leq \frac{VG^2}{\alpha} (1+\ln(t)) + \frac{G^2}{\alpha V}(1+\ln(t)) \big(\Phi'(Q(t))\big)^2, ~ \forall x^\star \in \mathcal{X}^\star, \forall t.
 \end{eqnarray}
 Finally, choosing $\Phi(\cdot)$ as the quadratic Lyapunov function, \emph{i.e.,} $\Phi(x) \equiv x^2,$ we arrive at the following result for strongly convex cost and convex constraint functions. 
 %the following theorem bounds the regret and the CCV for Algorithm \ref{coco_alg}.

 \begin{theorem} \label{str-cvx-bd}
 	% For COCO with $\alpha$-strongly convex cost and convex constraint functions, 
 	 For the COCO problem with adversarially chosen $\alpha$-strongly convex, $G$-Lipschitz cost functions and $G$-Lipschitz convex constraint functions,
 	 Algorithm \ref{coco_alg}, with $\beta=1, V=\frac{8G^2 \ln(Te)}{\alpha}, \Phi(x)= x^2,$ yields the following Regret and CCV bounds for any horizon length $T \geq 1:$
	%\begin{eqnarray*}
		\[ \textrm{Regret}_t(x^\star) \leq  \frac{G^2}{\alpha}\big(1+\ln(t)\big), ~ \textrm{CCV}_t= O\big(\sqrt{\frac{t \log T}{\alpha}}\big), \forall x^\star \in \mathcal{X}^\star, ~\forall t \in [T].\]
	%\end{eqnarray*}
Furthermore, if the worst-case regret is non-negative in some round $t$ (\emph{i.e.,} $\sup_{x^\star \in \mathcal{X}^\star}\textrm{Regret}_t(x^\star) \geq 0$), then the CCV can be further improved to $\textrm{CCV}_T= O(\frac{\log T}{\alpha})$ while keeping the regret bound the same. 
%(b) For a natural class of adversaries, called convex adversaries defined in Eqn.\ \eqref{jensenadv} in Appendix \ref{improved_rates}, we have $\mathbb{V}(t)= O(\frac{\log T}{\alpha})$ under some mild assumptions. See Theorem \ref{improved_violation_bd} for a precise statement.
	%In the above, the notation $\tilde{O}(\cdot)$ hides factors logarithmic in $T$.
 \end{theorem}
 %\end{framed}
 %\vspace{-0.1in}
 Please refer to Appendix \ref{str-cvx-pf} for the proof of Theorem \ref{str-cvx-bd}. 
 
 \textbf{Remarks:} The second part of the theorem is surprising because it says that when the regret is non-negative, a stronger logarithmic CCV bound holds for not necessarily strongly convex constraints. In Appendix \ref{improved_rates}, we give example of an interesting class of adversaries, called \emph{convex adversary}, for which the non-negative regret assumption holds true in the OCO setting.
 %Hence, the strong convexity of the cost functions leads to a sharper bound for CCV when the regret is non-negative. 
 
 \subsection{Lower Bounds} \label{lower_bound_section}
We now show that under Assumptions \ref{cvx}, \ref{bddness}, and \ref{feas-constr}, the regret and the CCV of any online policy for the COCO problem for $T$ rounds are both lower bounded by $\Omega(\sqrt{T})$ provided the problem is high-dimensional.  
Recall that if the constraint function $g_t= 0, \forall t$, then the COCO problem reduces to the standard OCO problem, and  $\Omega(\sqrt{T})$ is a well-known regret lower bound for OCO  \citep[Theorem 10]{hazan2022introduction}. In this case, we trivially have {$\textrm{CCV}= 0.$}  
The main challenge in proving a lower bound for COCO is \emph{simultaneously} bounding both the regret and CCV. Prior work does not give any simultaneous lower bounds since the standard adversarial inputs used to derive the lower bound of \citet{hazan2022introduction} do not satisfy the feasibility assumption (Assumption 3). We derive the lower bound by constructing a sequence of cost and constraint functions that satisfy Assumption \ref{feas-constr} in a $d$-dimensional Euclidean box of unit diameter.

%Note that the CCV lower bound does not follow from the corresponding regret lower bound for the OCO problem due to the additional Assumption \ref{feas-constr}, which puts an implicit constraint on the admissible constraint functions.

\begin{theorem}\label{thm:lbcoco}
Under Assumptions \ref{cvx}, \ref{bddness}, and \ref{feas-constr}, for any choice of the horizon length $T$ and online policy, there exists a problem instance with dimension $d \geq T$ where $\min (\textrm{Regret}_T, \textrm{CCV}_T) = \Omega(\sqrt{T}).$  
%the regret and CCV as defined in \eqref{intro-regret-def} and \eqref{intro-gen-oco-goal} with $k=1$ are both $\Omega(\sqrt{T})$.
\end{theorem}
In high-dimensional problems where $d \gg T,$ the above lower bound matches with the upper bound given in Theorem \ref{main_result}. The proof of Theorem \ref{thm:lbcoco} can be found in Appendix \ref{app:lbcoco}.

\section{The Online Constraint Satisfaction Problem (\textsc{OCS})} \label{simul_constr}
%\vspace{-0.2in}
%We begin our discourse with $\ocs,$ 
In this section, we study
a special case of the COCO problem, which involves only constraint functions and no cost functions. The OCS problem arises in many practical settings, including the multi-task learning problem (see Section \ref{mtl-ocs} in the Appendix for a brief discussion). In Section \ref{cbc} in the Appendix, we also establish a connection between the OCS problem and the well-studied Convex Body Chasing problem \citep{argue2019nearly}. The setup is similar to the COCO setting -- on every round $t\geq 1$, an online policy selects an action $x_t$ from a closed, bounded, and convex admissible set $\mathcal{X} \subseteq \mathbb{R}^d$. After observing the current action $x_t$, the adversary chooses $k$ constraints of the form $g_{t,i}(x) \leq 0, i\in [k],$ where each $g_{t,i}: \mathcal{X} \mapsto \mathbb{R} $ is a convex function. 
 Let $\mathcal{I}$ be any sub-interval of the horizon $[1,T].$ The cumulative constraint violation (CCV) $\mathbb{V}(T)$ for the \textsc{OCS} problem is defined as the maximum \emph{signed} cumulative constraint violation in any sub-interval:
 \begin{eqnarray} \label{violation-def1}
		\mathbb{V}(T) = \max_{i=1}^k \mathbb{V}_i(T), ~ \textrm{where} ~ \mathbb{V}_i(T) = \max_{\mathcal{I} \subseteq [1,T]}\sum_{t \in \mathcal{I}} g_{t,i}(x_t), ~ 1\leq i \leq k.
\end{eqnarray}
The objective is to design an online learning policy so that $\mathbb{V}(T)$ is as small as possible.
It is worth noting that in the $\ocs$ problem, we consider a soft constraint violation metric $\max_{\mathcal{I}}\sum_{t \in \mathcal{I}} g_{t,i}(x_t)$ instead of the hard violation metric $\sum_{t=1}^T (g_{t,i}(x_t))^+$ as in COCO. This allows for compensating the infeasibility on one round with strict feasibility on other rounds.
%, under the relaxed assumptions discussed below. %Essentially, this helps get 
%stronger CCV bounds than COCO. If a model requires $\sum_{t}\max\{g_{t,i}(x_t), 0\}$ to be the constraint violation cost, then results derived in Theorem \ref{gen-cvx-bd} and \ref{str-cvx-bd} for COCO will apply.
%For constraint functions that can take both positive and negative values, controlling the cumulative violations over all sub-intervals is stronger than controlling the cumulative violations over the full horizon. 
In contrast with the COCO setting, without Assumption \ref{feas-constr}, running a no-regret policy on the pointwise maximum of the constraint functions no longer works as the CCV of any fixed benchmark could grow linearly with 
%the horizon length 
$T$. In the OCS problem, we relax the feasibility assumption (Assumption \ref{feas-constr}), and consider the following two distinct alternatives instead. 
 \paragraph{\bf 1. $S$-feasibility:} Here, we assume that there is an admissible action $x^\star \in \mathcal{X}$ that satisfies the aggregate constraints over any interval of $S$ rounds. However, unlike \citet{georgios-cautious}, which also considers the same assumption, the value of the parameter $S$ is not necessarily known to the policy \emph{a priori}. Towards this end, we define the set of all $S$-feasible actions $\mathcal{X}_S$ as below: 
\begin{eqnarray} \label{extended-benchmark}
\mathcal{X}_S =\{x^\star \in \mathcal{X}: \sum_{\tau \in \mathcal{I}} g_{\tau,i}(x^\star) \leq 0, \forall \textrm{sub-intervals}~ \mathcal{I} \subseteq [1,T], ~|\mathcal{I}| = S, \forall i \in [k]\}. 
\end{eqnarray}
We now replace Assumption \ref{feas-constr} with the following weaker version:
\begin{assumption}[$S$-feasibility] \label{s-feas-assump}
	$\mathcal{X}_S \neq \emptyset$ for some $1\leq S \leq T.$
\end{assumption}
%In our analysis, we assume that $\mathcal{X}_S \neq \emptyset$ for some $S \in [T],$ which need not be known to the policy. 
Clearly, Assumption \ref{s-feas-assump} is weaker than Assumption \ref{feas-constr} as $\mathcal{X}^\star \subseteq \mathcal{X}_S, \forall S \geq 1.$ Note that even when the individual constraint functions satisfy $S$-feasibility, their pointwise maximum need not satisfy $S$-feasibility. Hence, unlike COCO under Assumption \ref{feas-constr}, this problem cannot be solved by simply running a no-regret policy on the pointwise maximum of the constraints.  

\paragraph{\bf 2. $P_T$-constrained adversary}
In this case, we drop any feasibility assumption altogether. As a consequence, any static admissible benchmark $x^\star \in \mathcal{X}$ also incurs a CCV. 
\begin{definition} \label{pt-feas-assump}
%We now consider a different relaxation to the instantaneous feasibility assumption where we now assume that 
An adversary is called $P_T$-constrained if its minimum static CCV is $P_TF$, \emph{i.e.,} 
$ \frac{1}{F} \min_{x^\star \in \mathcal{X}} \max_{\mathcal{I} \subseteq [T],i} \sum_{t \in \mathcal{I}} g_{t,i}(x^\star) = P_T$,  where $F$ is a normalizing factor denoting the maximum absolute value of the constraint functions within the compact admissible set $\mathcal{X}$. 
\end{definition}
As before, the value of $P_T$ is not necessarily known to the policy \emph{a priori}.
\subsection{Designing an \textsc{OCS} Policy with a Quadratic Lyapunov function} \label{meta-policy-ocs}
We define a process $\bm{Q}(t)=\big(Q_i(t), i \in [k]\big), t \geq 1$, which tracks the CCV:
\begin{eqnarray}\label{q-ev2}
	Q_i(t) = \big(Q_i(t-1)+ g_{t,i}(x_t)\big)^+, ~ Q_i(0)=0, ~t \geq 1, ~\forall i \in [k].
\end{eqnarray}
Notably, in contrast to COCO, we \emph{do not} clip the constraint functions in the above recursion. 
%where we have used the notation $y^+ \equiv \max(0,y).$
Expanding Eqn.\ \eqref{q-ev2}, which is also known as the queueing recursion or the \emph{Lindley process} \citep[pp. 92]{asmussen2003applied}, and using the definition in Eqn.\ \eqref{violation-def1}, we have the following relation for all $i \in [k]:$
\begin{eqnarray} \label{V-Q}
	\mathbb{V}_i(T) \equiv \max_{t=1}^T\max (0,\max_{\tau=0}^{t-1} \sum_{s=t-\tau}^t g_{s,i}(x_l))=\max_{t=1}^TQ_i(t). 
\end{eqnarray}
Equation \eqref{V-Q} indicates that to control the CCV \eqref{violation-def1}, it is sufficient to control the $\bm{Q}(t)$ process. Similar to the COCO problem, we combine the classic Lyapunov method with adaptive no-regret OCO policies to control the $\bm{Q}(t)$ process. 

%\section{An Online Meta-Policy} \label{policy}
%We will keep track of the cumulative constraint violation through the deterministic scalar process referred to as \emph{queue lengths} $\{Q(t)\}_{t \geq 1}$ that evolves as follows:
% \footnote{If the penalty function $\psi$ is also given to be non-negative, then the $(\cdot)^+$ operation in the definition of the recursion becomes superfluous.}:
% \begin{eqnarray} \label{q-ev}
% 	Q(t) = \big(Q(t-1)+ \psi(g_t(x_t))\big)^+, ~Q(0)=0. \end{eqnarray}
%\vspace{-0.1in}
\paragraph{A Quadratic Lyapunov function:}
 	 We consider the quadratic potential function %\begin{eqnarray*}\label{eq:potfunc}
$\Phi(\bm{Q}(t))\equiv \sum_{i=1}^k Q_i^2(t), t\geq 1.$
%\vspace{-0.1in}
%\end{eqnarray*} 
Since $((x)^+)^2=xx^+, \forall x\in \mathbb{R},$ from Eqn.\ \eqref{q-ev2}, we have 
%that for each $1\leq i\leq k:$
	 %\vspace{-0.1in}
 \begin{eqnarray} \label{q-bd2}
 	Q_i^2(t) &=& \big(Q_i(t-1)+ g_{t,i}(x_t)\big)Q_i(t) = Q_i(t-1)Q_i(t)+ Q_i(t) g_{t,i}(x_t), \nonumber \\
 	&\stackrel{(a)}{\leq}& \frac{1}{2}Q_i^2(t)+ \frac{1}{2}Q_i^2(t-1) + Q_i(t) g_{t,i}(x_t), ~ \forall i \in [k]. 
 \end{eqnarray}
 where in $(a)$, we have used the AM-GM inequality. Rearranging Eqn.\ \eqref{q-bd2},
% where $\max_{x \in \mathcal{X}} g_t^2(x)+ b_t^2 + 2|g_t(x)b_t| = \max_{x \in \mathcal{X}} (|g_t(x)|+|b_t|)^2\leq B.$ 
 %the \emph{one-step drift} 
 the change of the potential function $\Phi(\bm{Q}(t))$ on round $t$ can be upper bounded as follows
 	 %\vspace{-0.125in}
 \begin{eqnarray} \label{drift-bd}
 	\Phi(\bm{Q}(t))-\Phi(\bm{Q}(t-1)) = \sum_{i=1}^k \big(Q_i^2(t)-Q_i^2(t-1)\big)\leq 2\sum_{i=1}^k Q_i(t)g_{t,i}(x_t).
 \end{eqnarray}
 Similar to \eqref{surrogate_new}, we now define a surrogate cost function $\hat{f}_t:\mathcal{X} \mapsto \mathbb{R}$ as a linear combination of the constraint functions with the coefficients given by the vector $\bm{Q}(t)$, \emph{i.e.,}
 %\vspace{-0.175in}
 \begin{eqnarray} \label{surrogate-def}
 	\hat{f}_t(x) \equiv 2\sum_{i=1}^kQ_i(t) g_{t,i}(x).
 \end{eqnarray}
 %\vspace{-0.11in}
Clearly, the surrogate cost function $\hat{f}_t(\cdot)$ is convex since the coefficients $Q_i(t)$'s are non-negative and the constraint functions are convex. Our \textsc{OCS} policy, described below, simply runs a regret-minimizing adaptive OCO subroutine on the surrogate cost function sequence \eqref{surrogate-def}.

 %\begin{framed}
\textbf{The \textsc{OCS} policy (Algorithm \ref{ocs-policy}):} Pass the surrogate cost functions $\{\hat{f}_t\}_{t\geq 1}$ to the AdaGrad algorithm which enjoys a data-dependent regret as given in part 1 of Theorem \ref{data-dep-regret} in the Appendix (Eqn.\ \eqref{cvx-reg-bd}).
 %\end{framed}
% \textcolor{blue}{Explain what data-dependent bound means}.
 %Note that the convex cost function $\hat{f}_t: \mathcal{X} \mapsto \mathbb{R}$ defined in Eqn.\ \eqref{surrogate-def} is a legitimate input cost function to any OCO subroutine as all information required to compute the surrogate cost $\hat{f}_t$ (\emph{i.e.,} $Q(t)$ and $\{g_{t,i}\}_{i=1}^k$ is causally known at the end of round $t$. 
 %A complete pseudocode of the proposed meta-policy is given in.
 %The proposed meta-policy is summarized in Algorithm \ref{ocs-policy}. 
 %To clarify this point further, let us explicitly illustrate the order of events on round $t$. 

\begin{algorithm} 
\caption{Online Policy for \textsc{OCS}}
\label{ocs-policy}
\begin{algorithmic}[1]
\State \algorithmicrequire{ Sequence of convex constraint functions $\{g_{t,i}\}_{i\in [k], t\geq 1}$, a closed and convex admissible set $\mathcal{X}$ with a finite Euclidean diameter $D,$ $\mathcal{P}_\mathcal{X}(\cdot)=$ Euclidean projection operator on the set $\mathcal{X}$ }
\State \algorithmicensure{  Sequence of admissible actions $\{x_t\}_{t\geq 1}$}
\State{\bf{Initialization}:} Set $ x_1 \in \mathcal{X}$ arbitrarily, $Q_i(0)=0, ~\forall i \in [k].$ 
\ForEach {each round $t \geq 1$}
\State Play $x_t,$ observe the constraint functions $\{g_{t,i}\}_{i \in [k]}$ revealed by the adversary. 
\State [\textbf{Update $\bm{Q}(t)$}] $Q_i(t) = (Q_i(t-1)+ g_{t,i}(x_t))^+, i \in [k]$.
\State [\textbf{Compute a subgradient}] %Compute the gradient of the surrogate cost function 
$\nabla_t \equiv \nabla \hat{f}_t(x_t) = 2\sum_{i=1}^kQ_i(t) \nabla g_{t,i}(x_t).$
\State [\textbf{AdaGrad step}]  Compute the next action $x_{t+1} = \mathcal{P}_\mathcal{X}(x_t - \eta_t \nabla_t)$, where 
  % \begin{eqnarray*}
   $\eta_t =
   	\frac{\sqrt{2}D}{2\sqrt{\sum_{\tau=1}^{t} ||\nabla_\tau||_2^2}}.$
   %	\end{eqnarray*}
%\State [\textbf{OCO step}] Feed $\hat{f}_t(x)$ to the base OCO sub-routine $\Pi$, which outputs an action $x_{t+1} \in \mathcal{X}.$ \label{oco-step}
\EndForEach
\end{algorithmic}
\end{algorithm}
%Note that, unlike some of the previous work based on the Lyapunov drift approach \citep{neely2017online, yu2016low}, Algorithm \ref{ocs-policy} takes full advantage of the adaptive nature of the base OCO sub-routine by exploiting the fact that the adversary is allowed to choose the surrogate cost function $\hat{f}_t$ \emph{after} seeing the current action of the policy $x_t,$ which determines the coefficients $Q_i(t)$'s.

\subsection{Performance Bounds} 
\label{ext}

 \begin{theorem} \label{S-benchmark}
Under Assumptions \ref{cvx}, \ref{bddness}, and \ref{s-feas-assump}, 
%the OGD policy with adaptive step-sizes given in part 1 of Theorem \ref{data-dep-regret} as a sub-routine, 
Algorithm \ref{ocs-policy} achieves the following CCV bound for the OCS problem: 
 	 $\mathbb{V}(T)= O(\max(\sqrt{ST},S )).$
 \end{theorem}

 \begin{theorem} \label{P_T-benchmark}
 Under Assumptions \ref{cvx} and \ref{bddness}, 
% the OGD policy with adaptive step-sizes given in part 1 of Theorem \ref{data-dep-regret} as a sub-routine achieves the following CCV bound 
Algorithm \ref{ocs-policy} achieves the following CCV bound for the OCS problem
 for any $P_T$-constrained adversary as given in Definition \ref{pt-feas-assump}: 
 	 \[\mathbb{V}(T)= O(P_T^{\nicefrac{1}{3}}T^{\nicefrac{2}{3}})+O(\sqrt{T}).\]
 \end{theorem}
 Trivially, we have $S\leq T$ and  $P_T \leq T.$ In the non-trivial case where either $S$ or $P_T$ increases \emph{sub-linearly} with the horizon length $T$, the above theorems yield sublinear CCV bounds.
\section{Experiments: Credit Card Fraud Detection} \label{expts}
%\paragraph{Online anomaly detection:}
\begin{figure}[ht]
    \centering
    \begin{minipage}{0.49\textwidth}
        \centering
        \includegraphics[scale=0.4]{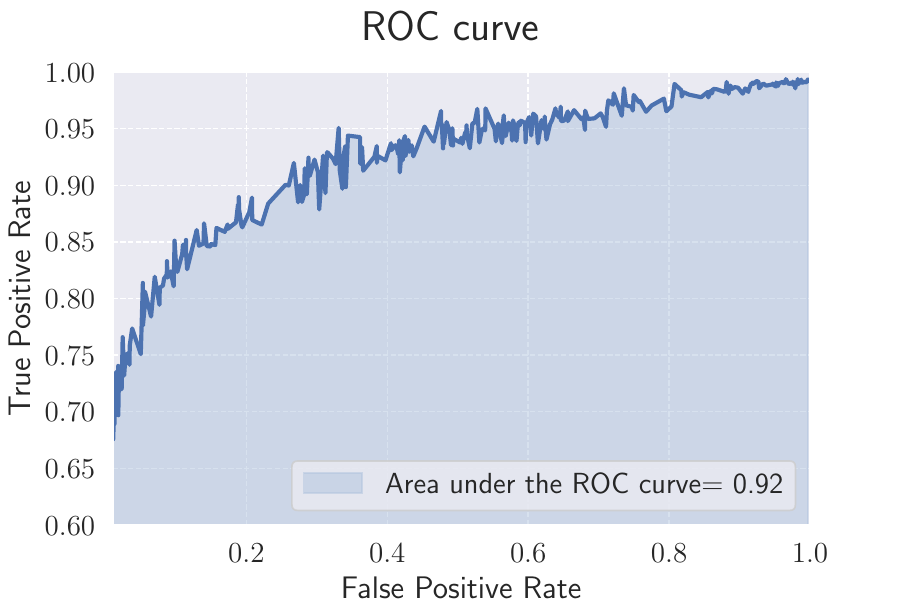}
        \caption{\small{ROC curve obtained by varying $\lambda$}}
        \label{fig:ROC}
    \end{minipage}
    \hfill
    \begin{minipage}{0.49\textwidth}
        \centering
                \includegraphics[scale=0.4]{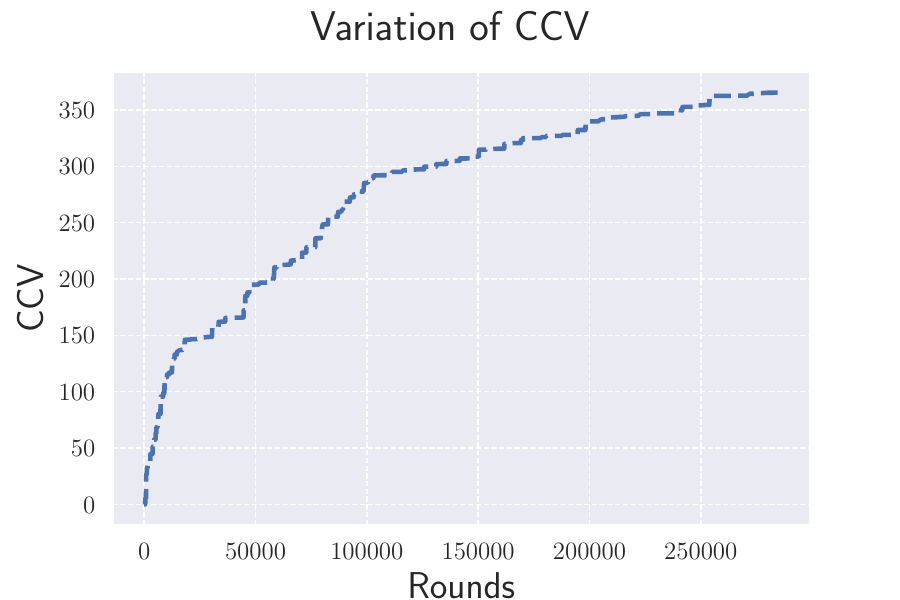}
        \caption{\small{Typical variation of the CCV with time}}
        \label{fig:ccv}
    \end{minipage}
\end{figure}
\paragraph{Classification with a highly imbalanced dataset:}
We first formulate the credit card fraud detection problem in the COCO framework. %We use Algorithm \ref{coco_alg} to train a neural network for detecting credit card fraud using a publicly available dataset. 
%We emphasize that in contrast with the standard resampling-based strategies for imbalanced classification, our policy is online. %The online operation is indispensable in the credit card fraud detection setting, where the algorithm needs to continuously learn in a dynamic environment and make real-time predictions.
Assume that we receive a sequence of $d$-dimensional feature vectors $\{z_t\}_{t \geq 1}$ and the corresponding binary labels $\{y_t\}_{t \geq 1}$ for a sequence of credit card transactions, where each transaction can either be legitimate (\texttt{label} $=0$) or fraudulent (\texttt{label} $=1$). The problem is to predict the label $\hat{y}_t$ for each transaction $z_t$ before its true label $y_t \in \{0,1\}$ is revealed. Typically, legitimate transactions outnumber fraudulent transactions by orders of magnitude. Since the goal is to detect any fraudulent transactions (even at the cost of a few false alarms), maximizing the classification accuracy alone is insufficient due to the significant class imbalance. We propose the following reformulation for this problem within the COCO framework. 
\vspace{-8pt}
\paragraph{Formulation:} Let $\hat{y}_t(z_t,x)$ be the likelihood of class $1$ for the feature $z_t,$ given by a parameterized model with parameter $x$. Hence, the log-likelihood $\mathcal{L}(t)$ of the data on round $t$ can be expressed as: 
\begin{align}
    \mathcal{L}(t) = y_t\log(\hat{y}_t(z_t, x))+ (1-y_t)\log(1 - \hat{y}_t(z_t, x)).
%    \begin{cases} 
%    \log(\hat{y}_t) & \text{if } y_t = 1 \\ 
%    \log(1 - \hat{y}_t) & \text{if } y_t = 0
%    \end{cases}
\end{align}
We train the model by maximizing the sum of log-likelihoods for legitimate transactions, subject to the constraint that all fraudulent transactions have a likelihood value close to $1$ (\emph{i.e.,} the sum of the log-likelihoods of the fraudulent transactions remains close to zero):
\begin{eqnarray} \label{prob1}
    \max_x \sum_{t=1}^T (1-y_t) \log(1-\hat{y}_t(z_t, x)),~~ \textrm{s.t.}~~\sum_{t=1}^T y_t \log(\hat{y}_t(z_t, x)) \geq 0.
\end{eqnarray}
%s.t. 
%\begin{eqnarray} \label{constr1}
%    \sum_{t=1}^T y_t \log(\hat{y}_t) \geq 0,
%\end{eqnarray}
%where the maximization is done over the parameters of the model. In other words,  %Problem \ref{prob1} can be rewritten as
%\begin{eqnarray} \label{main_prob}
%    \min \sum_{t=1}^T -(1-y_t) \log(1-\hat{y}_t) 
%\end{eqnarray}
%s.t. 
%\begin{eqnarray} \label{new-constr}
%    \sum_{t=1}^T -y_t (\log(\hat{y}_t)) \leq 0
%\end{eqnarray}
The above problem \eqref{prob1} can be immediately recognized to be an instance of COCO with the following cost and constraint functions:
\[ f_t(x) \equiv -(1-y_t) \log(1-\hat{y}_t(z_t,x)), ~~g_t(x)\equiv -y_t \log(\hat{y}_t(z_t,x)), ~t\geq 1.\]
In our experiments, we consider the common scenario in which the likelihoods are modeled by the output of a feedforward neural network. Note that the feasibility assumption (Assumption 3) is naturally satisfied as the overparameterized neural network models are known to perfectly fit the data \citep{belkin2019reconciling}. However, in this case, the functions $f_t$ and $g_t$ are generally non-convex. 
%Nevertheless, we find that Algorithm 1 performs excellently even when the convexity assumption does not hold. 
\vspace{-8pt}
\paragraph{Experiments:}

We experiment with a publicly available credit card transaction dataset \citep{dal2014learned}. This highly imbalanced dataset contains only $492$ frauds ($\sim 0.17\%$) out of $284,807$ reported transactions. 
%\paragraph{Dataset and Network Architecture:} 
Each data point has $D_{\textrm{in}}=30$ features and binary labels. We choose a simple network architecture with a single hidden layer containing $H=10$ hidden nodes and sigmoid non-linearities. Unlike previous algorithms, our algorithm is especially suitable for training neural network models as it only needs to compute the gradients (via backward pass) and evaluate the functions (via forward pass). Initially, all weights are independently sampled from a standard  normal distribution. The network is then trained  using Algorithm \ref{coco_alg} on a quad-core CPU with 8 GB RAM. The projection operation corresponds to $L_2$-normalization. The code has been  publicly released \citep{coco-code}.

%
%\begin{figure}
%    \centering
%    \includegraphics[width=0.7\linewidth]{./figures/ROC_plt.png}
%    \caption{Plot of the ROC curve}
%    \label{fig:ROC}
%\end{figure}
%
%\begin{figure}
%    \centering
%    \includegraphics[width=0.7\linewidth]{./figures/CCV_variation_Fraud_detection.png}
%    \caption{A plot of typical variation of CCV with time}
%    \label{fig:ccv}
%\end{figure}
\vspace{-8pt}
\paragraph{Results:}

Given the severe class imbalance, the area under the ROC curve, which plots the True Positive Rate (TPR) against the False Positive Rate (FPR), is an appropriate metric to evaluate any prediction algorithm for this problem. By varying the hyperparameter $\lambda$, we obtain the ROC curve shown in Figure \ref{fig:ROC}. The area under the ROC curve is computed to be $\approx 0.92$, which is an excellent score (cf. ideal score $=1.0$), notwithstanding the fact that, unlike the standard resampling-based techniques, the algorithm learns in an entirely online fashion starting from random initialization. Figure \ref{fig:ccv} illustrates the expected sublinear variation of CCV during one of the algorithm runs.
%\paragraph{References}
%
%[1] Dal Pozzolo, Andrea, Olivier Caelen, Yann-Ael Le Borgne, Serge Waterschoot, and Gianluca Bontempi. ``Learned lessons in credit card fraud detection from a practitioner perspective." Expert systems with applications 41, no. 10 (2014): 4915-4928.
\vspace{-0.15in}
\section{Conclusion} \label{conclusion}
\vspace{-0.15in}
\iffalse
An interesting open question is whether the assumption of the feasibility of the constraints on every slot can be suitably relaxed and still ensure sublinear regret and constraint violation penalties. Specifically, it would be interesting to extend our results to the $K$-benchmark case \citep{georgios-cautious} where the cumulative constraints hold for any consecutive $K$ intervals where $K$ is a sublinear function of the time-horizon $T$, \emph{i.e.,}
\begin{eqnarray*}
	\sum_{\tau \in \mathcal{I}}g_\tau (x^\star) \leq 0
\end{eqnarray*}
for all intervals $|\mathcal{I}| =K \leq \alpha(T)$ where $\alpha(\cdot)$ is a sublinear function. Secondly, proving joint lower bounds for static regret and constraint violation penalty would be interesting. Throughout this paper, we use the quadratic potential function to design our online policy. However, our regret decomposition result is general and could be used with any reasonable potential function. It would be interesting to see if improved performance guarantees can be established by choosing a different potential function.
\fi
In this paper, we proposed efficient online policies for the COCO problem with optimal performance bounds. We also derived sublinear CCV bounds for the OCS problem under a set of relaxed assumptions. Our analysis is streamlined, leveraging Lyapunov theory and adaptive regret bounds for the standard OCO problem. 
In the future, exploring dynamic regret bounds and a bandit extension of the COCO problem would be interesting.
%\iffalse
\section{Acknowledgement}
This work was supported by the Department of Atomic Energy, Government of India, under project no. RTI4001 and by a Google India faculty research award. The first author was also partially supported by a US-India NSF-DST collaborative grant coordinated by IDEAS-Technology Innovation Hub (TIH) at the Indian Statistical Institute, Kolkata. The authors gratefully acknowledge comments from the anonymous reviewers, which substantially improved the quality of the presentation. 
%
%\fi

%\bibliographystyle{plain}
%\clearpage
\bibliography{OCO.bib} 
\bibliographystyle{unsrtnat}
%\clearpage

%%%%%%%%%%%%%%%%%%%%%%%%%%%%%%%%%%%%%%%%%%%%%%%%%%%%%%%%%%%%

\appendix
\newpage
\section{Appendix} \label{appendix}
\subsection{Discussion on Assumptions \ref{cvx}, \ref{bddness} and \ref{feas-constr}}\label{app:assumptions}
Assumptions \ref{cvx} and \ref{bddness} are standard in the online learning literature. The feasibility assumption (Assumption \ref{feas-constr}) is analogous to the \emph{realizability} assumption in learning theory \citep{pmlr-v178-hopkins22a} and is commonly used in the COCO literature \citep{neely2017online, yu2016low,yuan2018online,yi2023distributed, georgios-cautious}. Assumption 3 requires the existence of a single admissible action $x^\star \in \mathcal{X}$ that satisfies the constraints in \emph{every} round. Consequently, all constraint functions are required to be non-positive over a non-empty common subset. This assumption is weakened in Section \ref{simul_constr}, Assumption \ref{s-feas-assump}, which only requires the existence of a fixed admissible action $x^\star$ that satisfies the constraints \emph{on average}. Specifically, Assumption \ref{s-feas-assump} requires that the sum of the constraint functions evaluated at some admissible $x^\star$ over any interval of length $S$ is non-positive. Notably, throughout the paper, we \emph{do not} assume Slater's condition as it does not hold in many problems of interest \citep{yu2016low}. As a result, unlike many previous works \citep{yu2017online}, our bounds are \emph{independent} of Slater's constant, which can be problem-dependent. Furthermore, we do not restrict the sign of either cost or constraint functions, allowing them to take both positive and negative values. 
%See the discussion on preprocessing the constraint functions in Section \ref{gen_oco}.
%Inspired by the Lyapunov method in the control theory, in the following, we propose an online meta-policy for the \textsc{OCS} problem and show that it yields optimal violation bounds. 
%Our main technical contribution is that while the classic works, such as \citet{neely2010stochastic}, use the Lyapunov theory in a stochastic setting; we adapt it to the adversarial setting by combining the Lyapunov method with the OCO framework.
%study the adversarial version of the problem through the lens of the OCO framework.
\subsection{Preliminaries on Online Convex Optimization (OCO)} \label{prelims}
 The standard OCO problem can be described as a repeated game between an online policy and an adversary \citep{hazan2022introduction}.
Let $\mathcal{X} \subseteq \mathbb{R}^d$ be a convex decision set, which we refer to as the \emph{admissible} set.
In each round $t\geq 1,$ an online policy selects an action $x_t \in \mathcal{X}.$ After the action $x_t$ is chosen, the adversary reveals a convex cost function $f_t : \mathcal{X} \mapsto \mathbb{R}$. 
%Since any convex function could be non-differentiable on a set of measures at most zero, without affecting the regret bounds, wherever convenient, we will assume the cost functions to be differentiable everywhere and replace gradients by subgradients at any point of non-differentiability \citep{hazan2007adaptive}. 
The goal of the online policy is to choose an admissible action sequence $\{x_t\}_{t\geq 1}$ so that its total cost over a horizon of length $T$ is not significantly larger than the total cost incurred by any fixed admissible action $x^\star \in \mathcal{X}$. More precisely, the objective is to 
minimize the static regret, defined as:
\begin{eqnarray} \label{regret-def}
	\textrm{Regret}_T \equiv \sup_{x^\star \in \mathcal{X}} \textrm{Regret}_T(x^\star), ~\textrm{where~}\textrm{Regret}_T(x^\star) \equiv \sum_{t=1}^T f_t(x_t) - \sum_{t=1}^T f_t(x^\star).
\end{eqnarray}

\begin{algorithm} 
\caption{Online Gradient Descent (OGD)}
\label{ogd-policy}
\begin{algorithmic}[1]
\State \algorithmicrequire{ Non-empty closed convex set $\mathcal{X} \subseteq \mathbb{R}^d$, sequence of convex cost functions $\{f_t\}_{t\geq 1},$ step sizes $\eta_1, \eta_2, \ldots, \eta_T >0,$ Euclidean projection operator $\mathcal{P}_\mathcal{X}(\cdot)$ onto the set $\mathcal{X}$}
\State{\textbf{Initialization}:} Set $x_1 \in \mathcal{X}$ arbitrarily
\ForEach {round $t\geq 1$}
\State Play $x_t$, observe $f_t$, incur a cost of $f_t(x_t)$.
\State Compute a (sub)gradient $\nabla_t \equiv \nabla f_t(x_t)$. %Define $G_t=||\nabla_t||_2.$
\State Update $x_{t+1}=\mathcal{P}_{\mathcal{X}}(x_t-\eta_t \nabla_t).$
\EndForEach 
\end{algorithmic}
\end{algorithm}

In a seminal paper, \citet{zinkevich2003online} showed that the online gradient descent policy, outlined in Algorithm \ref{ogd-policy}, run with an appropriately chosen constant step size sequence, achieves a sublinear regret bound $\textrm{Regret}_T = O(\sqrt{T})$ for Lipschitz-continuous convex cost functions. 
%In this paper, we are interested in stronger adaptive regret bounds where the bound is given in terms of the norm of the gradients and the strong-convexity parameters of the online cost functions. 
In Theorem \ref{data-dep-regret}, we recall two standard results on further refined data-dependent adaptive regret bounds achieved by the OGD policy with appropriately chosen adaptive step size sequences. 
%We will use the OGD policy with an appropriate step size sequence as a subroutine in our proposed online algorithm. 
%\edit{Take a look at \cite{yang2014regret} for variational bounds. Also, take a look at the case of exp-concave losses given in \cite{zhang2022simple}}.
\begin{theorem} \label{data-dep-regret}
Consider the generic OGD policy outlined in Algorithm \ref{ogd-policy}. 
%Depending on the step-size sequence, the OGD policy achieves the following data-dependent adaptive regret bounds. 
\begin{enumerate}
	%\item {\cite[Theorem 4.14]{orabona2019modern}} 
	\item {\citep{duchi2011adaptive}, \cite[Theorem 4.14]{orabona2019modern}} Let the cost functions $\{f_t\}_{t \geq 1}$ be convex and the step size sequence be adaptively chosen as $\eta_t= \frac{\sqrt{2}D}{2\sqrt{\sum_{\tau=1}^{t} G_\tau^2}}, t \geq 1,$ where $D$ is the Euclidean diameter of the admissible set $\mathcal{X}$ and $G_t=||\nabla f_t(x_t)||_2, t\geq 1.$ Then Algorithm \ref{ogd-policy} achieves the following regret bound: 
	\begin{eqnarray} \label{cvx-reg-bd}
			 \textrm{Regret}_T \leq \sqrt{2}D \sqrt{\sum_{t=1}^T G_t^2}.
	\end{eqnarray}
	The OGD policy with the above adaptive step-size sequence is known as (a variant of) the AdaGrad policy in the literature \citep{duchi2011adaptive}. 

	\item {\cite[Theorem 2.1]{hazan2007adaptive}} Let the cost functions $\{f_t\}_{t\geq 1}$ be strongly convex and let $H_t>0$ be the strong convexity parameter\footnote {The strong convexity of $f_t$ implies that $f_t(y)\geq f_t(x) + \langle \nabla f_t(x), y-x\rangle + \frac{H_t}{2}||x-y||^2, \forall x, y \in \mathcal{X}, \forall t.$} for the cost function $f_t$. Let the step size sequence be adaptively chosen as $\eta_t = \frac{1}{\sum_{s=1}^{t} H_s}, t \geq 1.$ Then Algorithm \ref{ogd-policy} achieves the following regret bound:
	\begin{eqnarray} \label{str-cvx-reg-bd}
		\textrm{Regret}_T \leq \frac{1}{2}\sum_{t=1}^T \frac{G_t^2}{\sum_{s=1}^t H_s}.
	\end{eqnarray} 
	\iffalse
	\item {\cite[Theorem 4.25]{orabona2019modern}, \cite[Theorem 2]{zhang2019adaptive}} Let the cost functions $\{f_t\}_{t \geq 1}$ be convex, non-negative, and $M$-smooth \footnote{$H$-smoothness means that $|| \nabla f_t(x)- \nabla f_t(y)||_2 \leq M||x-y||_2, \forall x,y\in \mathcal{X}, \forall t.$ }. Then the OGD policy with the same step-size sequence as in part 1 achieves the following regret bound:
	\begin{eqnarray*}
		\mathcal{R}_T(x^\star) \leq 4MD^2 + 4D\sqrt{M\sum_{t=1}^T f_t(x^\star)},
	\end{eqnarray*}
	where $x^\star \in \mathcal{X}$ is the static benchmark used in the definition of regret \eqref{regret-def}.
	\fi
	\end{enumerate}
\end{theorem}
%Note that the above adaptive bounds are mentioned for the simplicity of the regret expressions and the corresponding policy and for no other particular reason. 
Similar adaptive regret bounds are known for various other online learning policies as well. For structured domains, one can use other algorithms such as AdaFTRL \citep{orabona2018scale} which gives better regret bounds for high-dimensional problems. Furthermore, for problems with combinatorial structures, adaptive oracle-efficient algorithms, \emph{e.g.,} Follow-the-Perturbed-Leader (FTPL)-based policies, can be employed \citep[Theorem 11]{abernethy2014online}. 
Our proposed policies are agnostic to the specific online learning subroutine used for the surrogate OCO problem - what matters is that the subroutine provides adaptive regret bounds similar to \eqref{cvx-reg-bd} and \eqref{str-cvx-reg-bd}. This flexibility allows for an immediate extension of our algorithm to a wide range of settings, such as delayed feedback \citep{joulani2016delay} or combinatorial actions.

\subsection{Online Multi-task Learning as an Instance of the \ocs~ Problem} \label{mtl-ocs}
\begin{figure}[!h]
 \centering
 	\includegraphics[scale=0.45]{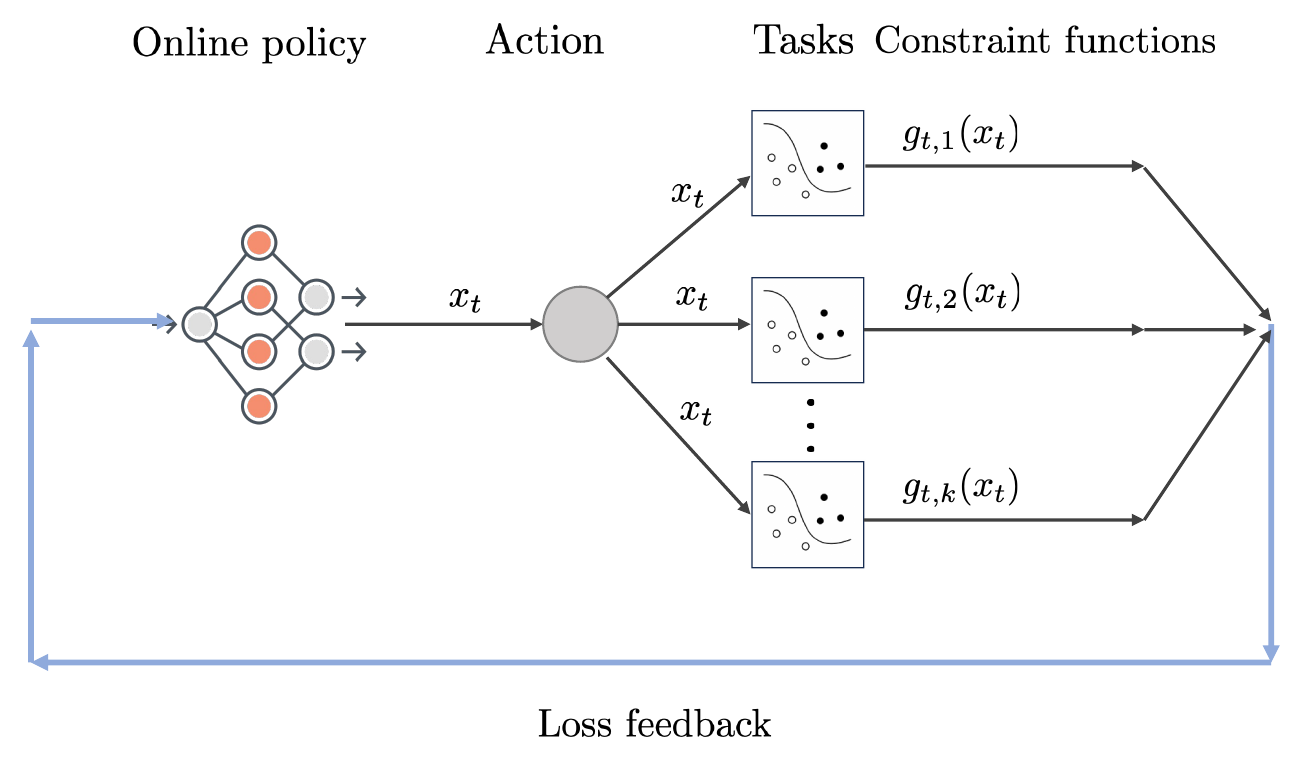}
  \put(-210,110){\small{Shared weights}}
 	\caption{A schematic for the online multi-task learning problem}
 	\label{multi-task}
 \end{figure}
%\paragraph{Example: Online Multi-task Learning:} 
Consider the problem of online multi-task learning where  a single model is trained to perform a number of related tasks \citep{ruder2017overview,dekel2006online, murugesan2016adaptive}. %The instances for each task may be chosen adversarially. 
See Figure \ref{multi-task} for a simplified schematic of the multi-task learning pipeline. In this setup, the action $x_t$ naturally corresponds to the shared weight vector that specifies the common model for all tasks. The loss function for the $j$\textsuperscript{th} task on round $t$ is given by the function $g_{t,j}(\cdot), j \in [k].$ A task is assumed to be satisfactorily completed (\emph{e.g.,} correct prediction in the case of classification problems) on any round if the corresponding loss is non-positive. As an example, using linear predictors for the binary classification problem, the requirement for the $j$\textsuperscript{th} task on round $t$ can be taken to be $g_{t,j}(x_t) \equiv \langle z_{t,j}, x_t\rangle \leq 0,$ where $z_{t,j}$ is the feature vector for the $j$\textsuperscript{th} task. The goal in multi-task learning is to sequentially update the shared weight vectors $\{x_t\}_{t=1}^T$ so that all tasks are successfully completed. Formally, we require that the maximum cumulative loss of each task over any sub-interval grows sub-linearly. Since the weight vector is shared across the tasks, the above goal would be impossible to achieve had the tasks not been related to each other \citep{ruder2017overview}. Theorem \ref{S-benchmark} and Theorem \ref{P_T-benchmark} give performance bounds for Algorithm \ref{ocs-policy} under different task-relatedness assumptions. 
\subsection{Proof of Theorem \ref{thm:lbcoco}}\label{app:lbcoco}
%\begin{proof}
    We prove the Theorem via constructing an explicit input sequence for which no online policy can have better than $\Omega(\sqrt{T})$ regret and CCV.
%\edit{We are proving only the violation bound and not the regret bound, right?}
\paragraph{Action space $\mathcal{X}$:} 
Let $d=T$. Let  $\mathcal{X}$  be the $d$-dimensional cuboid $0\le x_i \le \frac{1}{\sqrt{d}}, \ 1\le i\le d$. Clearly, the Euclidean diameter of $\mathcal{X}$ is $1$.

\paragraph{Input:} For each round we will only consider the case when only one constraint is revealed, \emph{i.e.}, $k=1$. On round $t=1, \dots, d$, choose the constraint $g_t$ to be $x_{t} \le \frac{1}{4\sqrt{d}}$ or $x_{t} \ge \frac{3}{4\sqrt{d}}$ with equal probability of $\frac{1}{2}$ for ${\bf x} = (x_1, \dots, x_d) \in \mathcal{X}$. Thus, at round $t$, only the $t\textsuperscript{th}$ dimension has an effective constraint.
If the chosen $g_t$ is $x_{t} \le \frac{1}{4\sqrt{d}}$ then pick $f_t= |x-\frac{1}{4\sqrt{d}}| $, otherwise pick  $f_t= |x-\frac{3}{4\sqrt{d}}|$.

For any online policy ${\cal A}$, the expected constraint violation at round $t$ is at least 
$\frac{1}{8\sqrt{d}}$. Thus, the overall expected constraint violation over rounds $t=1, \dots, d$ is at least 
$\frac{\sqrt{d}}{8}$. Moreover, the expected cost ${\mathbb E}[ f_t(x_t)]$ of ${\cal A}$ is at least $\frac{1}{8\sqrt{d}}$ for each $t=1, \dots, d$, and the overall cost ${\mathbb E}\big[\sum_{t=1}^T f_t(x_t)\big]$ is at least $\frac{\sqrt{d}}{8}$.

Recall that the choice of input has to satisfy Assumption \ref{feas-constr}, i.e., $\mathcal{X}^\star \neq \emptyset.$ We next demonstrate that for  the prescribed input $\exists \ {\bf x}^\star\in \mathcal{X}^\star$.
\paragraph{Choosing a feasible ${\bf x}^\star$:} When $g_t$ is such that the constraint is $x_{t} \le \frac{1}{4\sqrt{d}}$ choose ${\bf x}^\star \in \mathcal{X}$ such that $x^\star_t = \frac{1}{4\sqrt{d}}$ for $t=1,2,\dots, d$,  while if $g_t$ is such that $x_{t} \ge \frac{3}{4\sqrt{d}}$, then choose $x^\star_t = \frac{3}{4\sqrt{d}}$ for $t=1,2,\dots, d$. 
Thus, a single vector ${\bf x}^\star$ satisfies all the revealed constraints. Moreover, with this choice of ${\bf x}^\star$, the  overall cost of  ${\bf x}$,  $\sum_t f_t({\bf x}^\star)$, is $0$.

Since $d=T$, we get that for any online policy ${\cal A}$ its regret is at least $\Omega(\sqrt{T})$ and the cumulative constraint violation is $\Omega(\sqrt{T})$.$~~~~\blacksquare$

\subsection{Comparison with Previous Works}
\label{app:comparisonpolicies}
\subsubsection{ \cite{neely2017online}, \cite{yu2017online} and \cite{georgios-cautious}}
%At a high-level our proposed policy is similar to  \cite{neely2017online}, \cite{yu2017online} and \cite{georgios-cautious}.  
Policies  proposed by \cite{yu2017online} and \cite{georgios-cautious} are almost  identical to \cite{neely2017online}. The policy proposed in 
\cite{neely2017online}, however, is highly customized, does not fully exploit 
the best guarantees available for the standard OCO problem, and obtains sub-optimal performance bounds that depend inversely on Slater constant, which is assumed to be strictly positive. In a nutshell, \cite{neely2017online} choose the next action $x_{t+1}$ using the algorithm described below. For all rounds $t \geq 1, $ define the following evolution for $Q(t):$
 \begin{eqnarray} \label{q-ev}
 	Q(t) = (Q(t-1)+ g_t(x_t) + \nabla^T g_t(x_t)(x_t-x_{t-1}))^+, ~Q(0)=0. 
	\end{eqnarray}
	The next action is chosen by solving the following quadratic optimization problem: 
	$$x_{t+1} = \arg \min_{x\in {\mathcal X} } \big[\langle V \nabla^T f_{t}(x_{t}) + Q(t)\nabla g_t(x_t), x \rangle + \alpha ||x-x_{t-1}||^2\big],$$
	where $V$ and $\alpha$ are suitably chosen parameters. 
	
	In comparison, we have a different and simpler update rule:
	 \begin{eqnarray} \label{q-ev}
 	Q(t) = Q(t-1)+ (2GD)^{-1}(g_t(x_t))^+, ~Q(0)=0. \end{eqnarray}
	We then construct a convex surrogate function $\hat{f}_t(x) \equiv f_t(x) + \frac{1}{4GD\sqrt{T}}e^{\frac{Q(t)}{2\sqrt{T}}}(g_t(x))^+,$ whose gradient is then passed directly to the AdaGrad subroutine.
	\paragraph{Remarks:} We emphasize that Theorem \ref{main_result}, which shows that it is possible to simultaneously achieve $O(\sqrt{T})$ regret and $\tilde{O}(\sqrt{T})$ CCV in the convex setting without assuming Slater's condition, is highly surprising and unexpected. In fact, \citet[Section 4]{georgios-cautious} had previously commented that: 
	
	"\emph{$\ldots$ On the other hand, the point $O(\sqrt{T})$, $O(\sqrt{T})$ achieved by \citet{neely2017online} for $K = 1$ is not part of our achievable guarantees; we attribute this gap to the stricter Slater assumption studied by \citet{neely2017online}." }  \\\\
	Theorem \ref{main_result} squarely falsifies the last conjecture. 
 \subsubsection{ \cite{guo2022online}}
The policy in \cite{guo2022online} is a slightly modified form of the policy proposed in \citep{neely2017online}. In particular, it chooses the action $x_t$ by solving the following quadratic optimization problem over $\mathcal{X}:$
\begin{eqnarray*}
	x_t = \arg \min_{x\in {\mathcal X}} \big[ \langle \nabla f_{t-1}(x_{t-1}), x-x_{t-1} \rangle + Q(t-1) \gamma_{t-1} g_{t-1}^+(x) + \alpha_{t-1}|| x- x_{t-1}||^2 \big],
\end{eqnarray*} 
where the $Q$ variables are updated as follows:
\[ Q(t)= \max(Q(t-1)+ \gamma_{t-1} g_{t-1}^+(x_t), \eta_t).\]

\iffalse
based on the Lagrangian principle of solving constrained optimization problem, where the 
Lagrangian is 
$$ L (x,\lambda_t) = f_t(x) + \lambda_t g_t(x).$$
%Since $f_t$ and $g_t$ are revealed after action $x_t$ is chosen, 
The cost function $f_t$ is linearly approximated by 
$${\hat f}_t(x)  = f_{t-1}(x_{t-1}) + \langle \nabla f_{t}(x_{t-1}), x-x_{t-1}\rangle ,$$
and the constraint function $g_t$ is replaced by $\gamma_{t-1}g_{t-1}(x)^+$, $\lambda_t$ by $Q_t$ which is updated as  $Q(t) = \max(Q(t-1)+ \gamma_{t-1}g_{t-1}(x_{t-1})^+, \eta_t)$ and a quadratic regularization term $\alpha_{t-1} ||x-x_{t-1}||^2$ is added to the Lagrangian. 
\fi

Here $\alpha_t, \eta_t, \gamma_t$ are suitably chosen learning rate parameters. Essentially, this policy is trying to find the local optimum of an augmented Lagrangian under the online information model ($f_t$ and $g_t$ are revealed after action $x_t$ is chosen). Since their augmented Lagrangian involves the constraint function $g_{t-1},$ their policy needs to solve a full-fledged constrained convex optimization problem over the set $\mathcal{X}$ after having full access to the constraint function. In comparison, our policy, rather than using approximations to Lagrangian and adding regularizers, makes full use of the well-developed theory for OCO and uses first-order methods that need to compute only a gradient and perform one Euclidean projection on each round.
\subsubsection{\cite{jenatton2016adaptive}}
The policy proposed by \cite{jenatton2016adaptive} is based on the idea of primal-dual algorithm for optimizing the augmented Lagrangian 
$$ L_t (\lambda,x) = f_t(x) + \lambda g_t(x) - \frac{\theta_t}{2} \ \lambda^2,$$ where $\frac{\theta_t}{2} \lambda^2$ is the augmentation term. The primal variable $x_t$ and the dual variable $\lambda_t$ are updated by executing projected gradient descent and gradient ascent on the Lagrangian as follows:
$$x_{t+1} = \mathcal{P}_\mathcal{X}(x_t - \eta_t \nabla_x L_t (x_t, \lambda_t))$$ and 
$$\lambda_{t+1} = (\lambda_t + \mu_t \nabla_\lambda L_t (x_t, \lambda_t))^+,$$
where $\theta_t, \eta_t$, and $\mu_t$ are parameters to be chosen.

 \subsection{Proof of Theorem \ref{str-cvx-bd}} \label{str-cvx-pf}
  %\subsubsection{Preliminaries} \label{prelim}
  \paragraph{Bounding the CCV:} 
Choosing $\Phi(x)=x^2$ in Eqn.\ \eqref{Gronwall-ineq}, we have for any feasible $x^\star \in \mathcal{X}^\star:$ 
\begin{eqnarray} \label{q-reg-str-cvx-bd}
	Q^2(t) + V\textrm{Regret}_t(x^\star) \leq \frac{VG^2}{\alpha} (1+\ln(t)) + \frac{4G^2Q^2(t)\ln(Te)}{\alpha V},
\end{eqnarray}
where, on the last term in the RHS, we have used 
%the non-decreasing property of the queue lengths and 
the fact that $t\leq T$. Setting $V = \frac{8G^2 \ln(Te)}{\alpha},$ and transposing the last term on the RHS to the left, the above inequality yields
\begin{eqnarray} \label{Q-bd-str-cvx}
	Q^2(t) + 2V\textrm{Regret}_t(x^\star) \leq \frac{2VG^2}{\alpha} (1+\ln(t)).
\end{eqnarray}
%Since the (sub)-gradients of the cost function is assumed to be bounded, as before, the regret at any round is lower bounded as $\textrm{Regret}_t(x^\star) \geq -GDt.$ 
Since the cost functions are assumed to be $G$-Lipschitz (Assumption \ref{bddness}),
we trivially have $\textrm{Regret}_t(x^\star) = \sum_{t=1}^T (f_t(x_t)-f_t(x^\star)) \geq -GDt.$
Hence, from Eqn.\ \eqref{Q-bd-str-cvx}, we obtain
\begin{eqnarray*}
	Q^2(t) \leq 2VGDt + \frac{2VG^2}{\alpha} (1+\ln(t)) \implies Q(t) \stackrel{(a)}{\leq} 4G \sqrt{\frac{GD}{\alpha}t \ln(Te)} + \frac{4G^2 \ln(Te)}{\alpha}.
\end{eqnarray*}
where step (a), we have substituted $V = \frac{8G^2 \ln(Te)}{\alpha}.$ Hence, we have the following bound $\textrm{CCV}_t = O\big(\sqrt{\frac{t \log T}{\alpha}}\big).$

\paragraph{Bounding the regret:} Using the above choice for the parameter $V$ and the fact that $Q^2(t)\geq 0,$ from Eqn.\ \eqref{Q-bd-str-cvx}, we have
%in Eqn.\ \eqref{q-reg-str-cvx-bd}, we have 
\begin{eqnarray*}
	2V\textrm{Regret}_t(x^\star) \leq \frac{2VG^2}{\alpha} (1+\ln(t)). 
	%- \frac{Q^2(t)}{2} \leq \frac{VG^2}{\alpha} \ln(t).
\end{eqnarray*}
%where we have set $V = \frac{2G^2 \ln(Te)}{\alpha}.$
This leads to the following logarithmic bound for regret for any feasible $x^\star \in \mathcal{X}^\star:$ 
\[ \textrm{Regret}_t(x^\star) \leq \frac{G^2}{\alpha} (1+\ln(t)).~~~~\blacksquare \]

\paragraph{A sharper CCV bound under the non-negative regret assumption:} We now establish an improved CCV bound when the worst-case regret is non-negative on  some round $t \geq 1$. Let $\sup_{x^\star \in \mathcal{X}^\star}\textrm{Regret}_t(x^\star) \geq 0$ for some round $t \geq 1.$ Letting $V = \frac{8G^2 \ln(Te)}{\alpha}$ as above, from Eq.\ \eqref{Q-bd-str-cvx} we have 
\begin{eqnarray*}
	Q^2(t) \leq \frac{2VG^2}{\alpha} (1+\ln(t)) \implies Q(t) = O\big(\frac{\ln T}{\alpha}\big), t \in [T]. ~\blacksquare
\end{eqnarray*}
\paragraph{Comment:} From the above proof, it immediately follows that the same conclusion holds even under the weaker assumption of $-\textrm{Regret}_T= O(\frac{\log T}{\alpha}).$ 
\subsection{Adversaries Ensuring Non-negative Regret} \label{improved_rates}
%In Theorem \ref{gen-cvx-bd} and \ref{str-cvx-bd}, we showed that under the assumption of the non-negativity of the worst-case regret, the constraint violation bounds can be improved to $O(\sqrt{{T}})$ and $O(\ln T/\alpha)$ for convex and strongly-convex cost functions, respectively. In this section, we additionally show that the same improved bounds hold in the case of a time-invariant fixed constraint function and a certain class of worst-case adversaries, called \emph{convex adversary}, defined next. COCO with time-invariant constraints has been studied in the literature where the main objective is to design gradient-based first-order policies that avoid the costly projection step on the constraint set \citep{jenatton2016adaptive, yuan2018online}. 

\paragraph{Convex adversary:} An adversary is called \emph{convex} if for any sequence of  action sequence $\{x_t\}_{t=1}^T,$ the adversary chooses the cost function sequence $\{f_t\}_{t=1}^T$ such that for any $T \geq 1,$ we have
\begin{eqnarray} \label{jensenadv}
	\sum_{t=1}^T f_t(x_t) \geq \sum_{t=1}^T f_t(\bar{x}_T),
\end{eqnarray}     
where $\bar{x}_T \equiv \frac{1}{T}\sum_{t=1}^T x_t.$ Hence, by definition, a convex adversary guarantees a non-negative regret with respect to the average action $\bar{x}_T$ for all rounds. In the following, we give two examples of convex adversaries.

 \paragraph{1. Fixed adversary:} An adversary which always selects a fixed convex function $f$ on all rounds is a convex adversary. In this case, Eqn.\ \eqref{jensenadv} holds due to the Jensen's inequality. 

\paragraph{2. Minimax adversary:} Let $\mathcal{F}$ denote an arbitrary non-empty set of convex functions defined on the admissible set $\mathcal{X}$. Consider an adversary $\mathcal{M}$, which, upon seeing the selected action $x_t$, chooses the worst cost function $f_t$ from the set $\mathcal{F}$ on round $t:$ 
\[f_t \in \arg\max_{f\in \mathcal{F}} f(x_t). \]
We now show that $\mathcal{M}$ is a convex adversary. By definition, for any round $\tau \in [T],$ we have 
\[ f_\tau(x_\tau) \geq f_t(x_\tau) \implies f_\tau(x_\tau) \geq \frac{1}{T} \sum_{t=1}^T f_{t}(x_\tau). \]
Summing up the above inequalities for each $\tau \in [T],$ we have 
\begin{eqnarray}\label{conv-adv-def}
 \sum_{\tau=1}^T f_\tau(x_\tau) \geq \sum_{t=1}^T \frac{1}{T}\sum_{\tau=1}^T f_t(x_{\tau}) \stackrel{\textrm{(a)}}{\geq} \sum_{t=1}^T f_t(\bar{x}_T),
 \end{eqnarray}
where inequality (a) follows upon applying Jensen's inequality to each cost function. 
Eqn.\ \eqref{conv-adv-def} shows that $\mathcal{M}$ is a convex adversary. 

P.S. It can be easily seen that Fixed adversary is a special case of Minimax adversary where $\mathcal{F}=\{f\}.$

\subsection{Connection Between \textsc{OCS} and the Convex Body Chasing Problem}  \label{cbc}
A well-studied problem related to the \textsc{OCS} problem is the
{\it nested convex body chasing (NCBC)} problem \citep{bansa2018nested,argue2019nearly,bubeck2020chasing}, 
where at each round $t$, a convex set $\chi_t \subseteq \chi$ is revealed such that 
$\chi_t\subseteq \chi_{t-1}$, where  $\chi_0=\chi \subseteq {\mathbb R}^d$ is a convex, compact, and bounded set. 
The objective is to choose  $x_t \in \chi_t$ so as to minimize the total movement cost across rounds
$C =   \sum_{t=1}^T  ||x_t - x_{t-1}||_2,$
where $x_0 \in \chi$ is some fixed action.
In NCBC, action $x_t$ is chosen \emph{after} the set $\chi_t$ is revealed. This is in contrast to the \textsc{OCS} problem, where $x_t$ must be chosen \emph{before} the constraints $g_{t,i}$'s are revealed at round $t$. Moreover, note that the nested condition $\chi_t \subseteq \chi_{t-1}$ is stricter than Assumption \ref{feas-constr}, which is applicable to the \textsc{OCS} problem.
However, as we show next, a feasible algorithm for NCBC also provides an upper bound on the CCV of the \textsc{OCS} problem under Assumption \ref{feas-constr}.

In this reduction, we define $\chi_t $ as the intersection of the first $kt$ convex constraints $g_{\tau,i} \leq 0, 1\leq \tau \leq t, i\in [k],$ revealed up to round $t$ for the \textsc{OCS} problem. It is easy to see that $\chi_t$ is convex and $\chi_t \subseteq \chi_{t-1}, \forall t.$
Let $x_t$ be the action chosen by an algorithm $\cal A$ for the NCBC problem after the set $\chi_t$ is revealed. Note that $\chi_t \neq \emptyset,$ thanks to Assumption \ref{feas-constr}. We now choose $y_{t} := x_{t-1}$ as the action for the \textsc{OCS} problem on round $t$, ensuring that action $y_t$ is chosen before the set $ \chi_t$ is revealed.
The resulting $i^{th}$ constraint violation for the \textsc{OCS} problem at round $t$ is given by 
\[
	g_{t,i}(y_{t}) \stackrel{(a)}\le g_{t,i}(y_{t}) - g_{t,i}(y_{t+1}) \le G ||y_{t} - y_{t+1}||,
\]
where $(a)$ follows from the feasibility of $\cal A$ for NCBC, $y_{t+1}= x_{t} \in \chi_{t}$ and hence $g_{t,i}(y_{t+1}) \leq 0$. Summing across rounds $t=1, \dots, T$, and taking the $\max$ over  all the $k$ constraints, we get that the CCV using $\cal A$ for the \textsc{OCS} is upper bounded by $ \sum_{t=2}^T G ||y_{t} - y_{t+1}|| \le \sum_{t=2}^T G ||x_{t-1} - x_{t}|| \le G \cdot C_{\cal A},$
where $C_{\cal A}$ is the movement cost of $\cal A$ for the NCBC problem.

From prior work \cite{bansa2018nested,argue2019nearly,bubeck2020chasing}, it is known that for NCBC, a Steiner point-based algorithm that chooses $x_t$ as the Steiner point of $\chi_t$ can achieve
$C_{\cal A} = O(\sqrt{d \log d})$, where $\chi \subset {\mathbb R}^d$. Thus, the Steiner point-based algorithm (even though computationally intensive) provides an $O(\sqrt{d \log d})$ constraint violation for the 
\textsc{OCS} as well. However, this result is effective for problems where  $\sqrt{d \log d} = o(T).$ Our result efficiently overcomes this hurdle and provides a bound under weaker feasibility assumptions even beyond $\sqrt{d \log d} = o(T)$ -- a setting that is better motivated in practice for modern deep learning applications which are characteristically high-dimensional.

%\section{Generalizing the \ocs ~problem with the $S$-feasibility assumption}
%\input{fast_rates}

\subsection{Proof of Theorem \ref{S-benchmark}} \label{S-benchmark-pf}
\label{ext}

\paragraph{Generalized regret decomposition:} Fix any $S$-feasible benchmark $x^\star \in \mathcal{X}_S,$ as given by Eqn.\ \eqref{extended-benchmark}. Then, from Eqn.\ \eqref{drift-bd}, we have 
\begin{eqnarray*}
	\Phi(\tau)- \Phi(\tau-1) &\leq& 2 \sum_{i=1}^k Q_i(\tau)g_{\tau, i}(x_\tau) \\
	&=& 2 \sum_{i=1}^k Q_i(\tau)\big(g_{\tau, i}(x_\tau)-g_{\tau, i}(x^\star)\big) + 2\sum_{i=1}^k Q_i(\tau)g_{\tau, i}(x^\star)\\
	&=& \hat{f}_\tau (x_\tau) - \hat{f}_\tau(x^\star)  +  2\sum_{i=1}^k Q_i(\tau)g_{\tau, i}(x^\star). 
\end{eqnarray*} 
 Summing up the above inequalities from $\tau = 1$ to $\tau=t,$ we have
 \begin{eqnarray} \label{new-reg-decomp}
 	\sum_{i=1}^k Q_i^2(t) =\Phi(t) \leq \textrm{Regret}'_t(x^\star) + 2 \sum_{i=1}^k\sum_{\tau=1}^t Q_i(\tau) g_{\tau, i}(x^\star),
 \end{eqnarray}
 where $\textrm{Regret}'(\cdot)$ refers to the regret of the surrogate costs as before. 
 We now bound the last term by making use of the $S$-feasibility of the action $x^\star$ as given by Eqn.\ \eqref{extended-benchmark}.
 Let us now divide the entire interval $[1,t]$ into disjoint and consecutive sub-intervals $\{\mathcal{I}_j\}_{j=1}^{\lceil t/S \rceil},$ each of length $S$ (except the last interval which could be of a smaller length). %Let $Q^\star_i(j) = \max_{\tau \in \mathcal{I}_j}Q_i(\tau)$ be the maximum queue
 Let $Q^\star_i(j)$ be the value of the variable $Q_i(\cdot)$ at the beginning of the $j$\textsuperscript{th} interval. We have
 \begin{eqnarray} \label{S-bd1}
 	\sum_{\tau=1}^t Q_i(\tau)g_{\tau, i}(x^\star) = \sum_{j=1}^{\lceil t/S \rceil} \sum_{\tau \in \mathcal{I}_j}\big(Q_i(\tau)-Q_i^\star(j)\big)g_{\tau, i}(x^\star) + \sum_{j=1}^{\lceil t/S \rceil}Q_i^\star(j) \sum_{\tau \in \mathcal{I}_j}g_{\tau, i}(x^\star) .   
 \end{eqnarray}
 %Let $g_{t,i}(x) \leq F, \forall x \in \mathcal{X}, t, i.$ 
% From the Lipschitzness assumption, we have $g_{t,i}(x) \leq GD \equiv F ~(\textrm{say}), \forall x \in \mathcal{X}, t, i.$ 
Using the boundedness assumption, let $g_{t,i}(x) \leq F, \forall x \in \mathcal{X}, t, i.$
 Using the Lipschitzness property of the queueing dynamics \eqref{q-ev2} with respect to time, we have 
 \begin{eqnarray*}
 	\max_{\tau \in \mathcal{I}_j} |Q_i(\tau)-Q_i^\star(j)| \leq F(S-1).
 \end{eqnarray*}
 Substituting the above bound into Eqn.\ \eqref{S-bd1}, we obtain 
 \begin{eqnarray} \label{new-Q-S}
 	\sum_{\tau=1}^t Q_i(\tau)g_{\tau, i}(x^\star)  \leq \big(1+\frac{t}{S}\big) F^2S(S-1) + F(S-1)(Q_i(t)+F(S-1)), 
 \end{eqnarray}
 where in the last term, we have used the $S$-feasibility of the action $x^\star$ in all intervals, except possibly the last interval. 
 %Clearly, when $S=1$, the RHS of the above bound becomes zero, and we recover Eqn.\ \eqref{q-regret-reln}. 
 Substituting the bound \eqref{new-Q-S} into Eqn.\ \eqref{new-reg-decomp}, we arrive at the following extended regret decomposition inequality:
 \begin{eqnarray} \label{gen-reg-decomp2}
 	\sum_{i=1}^k Q_i^2(t) &\leq& \textrm{Regret}'_t(x^\star) +  2kF^2S t + 2FS\sum_{i=1}^k Q_i(t) + 4F^2S^2k.
%&\leq & \textrm{Regret}'_t(x^\star) + 6kF^2S t + 2FS \sqrt{k} \sqrt{\sum_{i=1}^k Q_i^2(t)}.
 \end{eqnarray}
Eqn.\ \eqref{gen-reg-decomp2} leads to the following bound on the cumulative constraint violation.
% 
% \begin{theorem} \label{S-benchmark}
%Using the OGD policy with adaptive step-sizes given in part 1 of Theorem \ref{data-dep-regret} as a sub-routine, Algorithm \ref{ocs-policy} achieves the following CCV bound with the $S$-feasibility assumption (Assumption \ref{s-feas-assump}) for convex constraints: 
% 	 \[\max_{i=1}^k\mathbb{V}_i(T)= O(\max(\sqrt{ST},S )).\]
% \end{theorem}
% See below for the proof of the result.
%
%\paragraph{Remarks:} Recall that our proof of the $O(\sqrt{T})$ regret bound for the COCO problem with the $1$-benchmark in Theorem \ref{gen-cvx-bd} crucially uses the non-negativity of the pre-processed constraint functions. However, with $S$-feasible benchmarks, pre-processing by clipping the constraints does not work as then the positive violations can not be cancelled with a strictly feasible violation on a different round. We leave the problem of obtaining an optimal $O(\sqrt{T})$ regret bound for Algorithm \ref{g-oco-policy} for the COCO problem with the $S$-feasibility assumption as an open problem. 
%It is not clear how to  
%which is better than $O(T^{1-\epsilon/4})$ constraint violation bound obtained by \citet{georgios-cautious}.

%\textbf{Note:} This extension is meaningful only for the convex case. Any strongly-convex function 
%\vspace{5pt}
%\hrule 
%\textbf{Note:} \footnote{This extension is meaningful only when the range of the constraint functions includes both positive and negative values. For non-negative constraints, clearly $\mathcal{X}_S = \mathcal{X}_1 \forall S\geq 1.$}\\
%\hrule

\subsubsection{CCV Bound} 
%\paragraph{Constraint violation for convex constraints:}
 We now apply the generalized regret decomposition bound given in \eqref{gen-reg-decomp2} to the case of convex constraint functions. Substituting the regret bound \eqref{cvx-reg-bd} of the AdaGrad policy into Eqn.\  \eqref{gen-reg-decomp2}, we have 
\begin{eqnarray*}
		\sum_{i=1}^k Q_i^2(t) \leq c_1 \sqrt{\sum_{\tau=1}^t \big(\sum_{i=1}^kQ_i^2(\tau)\big)}+ c_2 S t +   c_3S\sum_{i=1}^k Q_i(t)+ c_4S^2
\end{eqnarray*}
where the constants $c_1 \equiv O(GD \sqrt{k}),c_2=O(kF^2), c_3=O(F), c_4=O(kF^2) $ are problem-specific parameters that depend on the bounds on the gradients and the maximum value of the constraint functions, the number of constraints,  and the diameter of the admissible set. Defining $Q^2(t) \equiv \sum_i Q_i^2(t),$ we obtain:
\begin{eqnarray*}
	Q^2(t) \leq c_1 \sqrt{\sum_{\tau=1}^t Q^2(\tau)} + c_2St + c_3 S \sum_{i=1}^k Q_i(t) + c_4S^2.
\end{eqnarray*}
Since $Q_i(t) \leq Ft, \forall i,$ the above inequality can be simplified to 
\begin{eqnarray} \label{simplified-q-bd}
	Q^2(t) \leq c_1 \sqrt{\sum_{\tau=1}^t Q^2(\tau)} +  c_2'St + c_4S^2, ~ \forall t\geq 1,
\end{eqnarray}
where we have defined $c_2'\equiv c_3kF+c_2.$
To solve the above system of inequalities, note that for each $1 \leq \tau \leq t,$ we have
\begin{eqnarray*}
	Q^2(\tau) \leq c_1 \sqrt{\sum_{\tau=1}^t Q^2(\tau)} +  c_2'St + c_4S^2.
\end{eqnarray*}
Summing up the above inequalities for $1\leq \tau \leq t$ and defining $Z_t \equiv \sqrt{\sum_{\tau=1}^t Q^2(\tau)},$ we obtain
\begin{eqnarray*}
	Z^2_t &\leq&  c_1t Z_t + c_2'St^2 + c_4S^2t \\
	\emph{i.e.,}~Z_t^2 &\leq & 3 \max(c_1t Z_t, c_2'St^2, c_4S^2t) \\
	\emph{i.e.,}~ Z_t &=& O(\max(t, t\sqrt{S}, S \sqrt{t})).  
\end{eqnarray*}
Substituting the above bound for $Z(t)$ in Eqn.\  \eqref{simplified-q-bd}, we have for any $t\geq 1$:
\begin{eqnarray*}
Q^2(t) &=& O(\max(Z_t, St, S^2)) \\
\emph{i.e.,}~ Q(t) &=& O(\max(\sqrt{Z_t}, \sqrt{St}, S))\\
\textrm{Hence,}~~ Q_i(t) \leq Q(t) &=& O(\max(\sqrt{t},  \sqrt{t}S^{1/4}, \sqrt{S} t^{1/4}, \sqrt{St}, S))\\
& =& O(\max(\sqrt{St},S )), ~\forall i\in [k]. 
\end{eqnarray*}
The final result follows upon appealing to Eqn.\ \eqref{V-Q}.$~~~~\blacksquare$
%In the special case $S=T^{1-\epsilon}, 0<\epsilon<1,$ the above bound yields $Q_i(t) = O(T^{1-\epsilon/2}), \forall t\geq 1.$ 
 %Hence, by setting $V=O(T),$ we get the optimal logarithmic regret for the strongly convex losses but it leads to a trivial linear constraint violation. However, by taking a smaller $V,$ we obtain a both sublinear regret and constraint violation penalty. In general, the following regret $R_T$ and violation penalty $\mathbb{V}_T$ profile is feasible (up to logarithmic factors in $T$): 
 %\begin{eqnarray*}
% 	(R_T, \mathbb{V}_T) = \tilde{O}(\frac{T}{V}, \sqrt{VT}).
% \end{eqnarray*}
% 

\subsection{Proof of Theorem \ref{P_T-benchmark}} \label{P_T_constrained}
We will use a similar line of arguments used in the analysis of an $S$-constrained adversary for a suitable value of $S$ to be determined later. We start from Eqn.\ \eqref{S-bd1}, which holds for any value of the sub-interval length $S \geq 1$ and any arbitrary adversary. Furthermore, from the definition of a $P_T$-constrained adversary, we know that there exists a benchmark $x^\star \in \mathcal{X}$ such that for any interval $\mathcal{I}_j$ and any $i \in [k],$ we have:
\[ \sum_{\tau \in \mathcal{I}_j} g_{\tau, i} (x^\star) \leq P_TF, \]
where $F$ is the maximum absolute value of the constraint functions as given in Definition \ref{pt-feas-assump}.
Hence, 
\begin{eqnarray*}
	\sum_{j=1}^{\lceil t/S \rceil}Q_i^\star(j) \sum_{\tau \in \mathcal{I}_j}g_{\tau, i}(x^\star) &\leq& P_TF \sum_{j=1}^{\lceil t/S \rceil}Q_i^\star(j) \\
	&\leq & \frac{P_TF}{S} \sum_{j=1}^{\lceil t/S \rceil} \sum_{\tau \in \mathcal{I}_j}\big(Q_i^\star(j)-Q_i(\tau)\big) + \frac{P_TF}{S}\sum_{\tau=1}^t Q_i(\tau).
\end{eqnarray*}
Hence, from Eqn.\ \eqref{S-bd1}, we have that 
\begin{eqnarray*}
	 	\sum_{\tau=1}^t Q_i(\tau)g_{\tau, i}(x^\star)  &\leq& \big(1+\frac{t}{S}\big) F^2S(S-1) + (1+\frac{t}{S})P_TF^2(S-1) + \frac{P_TF}{S}\sum_{\tau=1}^t Q_i(\tau) \\
	 	&\leq & F^2(S+P_T)(S+t)+ \frac{P_TF}{S}\sum_{\tau=1}^t Q_i(\tau).
\end{eqnarray*}
Substituting the above bound into Eqn.\ \eqref{new-reg-decomp}, we have that 
\begin{eqnarray*}
	\sum_{i=1}^k Q_i^2(t) \leq \textrm{Regret}_t'(x^\star) + 2kF^2(S+P_T)(S+t)+ \frac{2P_TF}{S}\sum_{\tau=1}^t \sum_{i=1}^k Q_i(\tau).
\end{eqnarray*}
Plugging in the regret bound of the AdaGrad policy for the surrogate cost functions, the above equation yields
\begin{eqnarray} \label{p_t-bd-eq}
		\sum_{i=1}^k Q_i^2(t) \leq GD \sqrt{2k}\sqrt{\sum_{\tau=1}^t \big(\sum_{i=1}^kQ_i^2(\tau)\big)}+ 2kF^2(S+P_T)(S+t)+ \frac{2P_TF}{S}\sum_{\tau=1}^t \sum_{i=1}^k Q_i(\tau).
\end{eqnarray} 
Using Cauchy-Schwarz inequality, the last term of the above inequality can be upper bounded by \[\frac{2P_TF \sqrt{kt}}{S} \sqrt{\sum_{\tau=1}^t \big(\sum_{i=1}^kQ_i^2(\tau)\big)}.\]
Hence, we have the following inequality which holds for any $1\leq S \leq t$ and $1\leq \tau \leq t:$
\begin{eqnarray} \label{p_t-bd-eq2}
		\sum_{i=1}^k Q_i^2(\tau) \leq \bigg(GD \sqrt{2k}+ \frac{2P_TF \sqrt{kt}}{S}\bigg)\sqrt{\sum_{\tau=1}^t \big(\sum_{i=1}^kQ_i^2(\tau)\big)}+ 2kF^2(S+P_T)(S+t).
\end{eqnarray} 
Summing up the above inequalities for $1\leq \tau \leq t$ and defining $Z_t^2 \equiv \sum_{\tau=1}^t \sum_{i=1}^kQ_i^2(\tau),$ we have:
\begin{eqnarray*}
	Z_t^2 &\leq& \bigg(GD \sqrt{2k}+ \frac{2P_TF \sqrt{kt}}{S}\bigg)tZ_t + 2kF^2t(S+P_T)(S+t) \\
	&\leq& 2 \max \bigg(\big(GD \sqrt{2k}+ \frac{2P_TF \sqrt{kt}}{S}\big)tZ_t, 2kF^2t(S+P_T)(S+t)\bigg). 
\end{eqnarray*}
The above inequality implies that 
\begin{eqnarray}
	Z_t &\leq& 2 \max \bigg(\big(GD \sqrt{2k}+ \frac{2P_TF \sqrt{kt}}{S}\big)t, F\sqrt{kt(S+P_T)(S+t)}\bigg) \nonumber \\
	&\leq & 2 \max \bigg(\big(GD \sqrt{2k}+ \frac{2P_TF \sqrt{kT}}{S}\big)T, FT\sqrt{2k(S+P_T)}\bigg), \label{Z-t-bd2}
\end{eqnarray}
where in the last step, we have used the fact that $t \leq T$ and $S \leq T$. Now, let us choose $S\equiv  P_T^{\nicefrac{2}{3}}T^{\nicefrac{1}{3}}$. With the above choice of $S$, from the above inequality, we have the following bound for $Z_t:$
\begin{eqnarray*}
	Z_t \leq 2 \max \bigg(\big(GD \sqrt{2k}+ 2F\sqrt{k}P_T^{\nicefrac{1}{3}}T^{\nicefrac{1}{6}}\big)T, 2F\sqrt{k}P_T^{\nicefrac{1}{3}}T^{\nicefrac{7}{6}}\bigg) = O(P_T^{\nicefrac{1}{3}}T^{\nicefrac{7}{6}})+O(T),
\end{eqnarray*}
where we have used the fact that $P_T \leq T.$
Substituting the above bound in \eqref{p_t-bd-eq2}, we have for any $1\leq i\leq k$ and any $t \leq T:$
\begin{eqnarray*}
	\sum_{i=1}^k Q_i^2(t) \stackrel{(a)}{=} O(P_T^{\nicefrac{2}{3}}T^{\nicefrac{4}{3}})+ O(T)+ O(P_T^{\nicefrac{2}{3}}T^{\nicefrac{4}{3}}) = O(P_T^{\nicefrac{2}{3}}T^{\nicefrac{4}{3}}) +O(T),
\end{eqnarray*}
where in (a), we have used the fact that $T\geq S\geq P_T$ in bounding the last term.
Hence, we have the following upper bound on the queue lengths for any $1\leq t \leq T$
\begin{eqnarray*}
||\bm{Q}(t)||_\infty \leq ||\bm{Q}(t)||_2 = O(P_T^{\nicefrac{1}{3}}T^{\nicefrac{2}{3}})+O(\sqrt{T}).
\end{eqnarray*}
The final result follows upon appealing to the relation \eqref{V-Q}. $~~~~\blacksquare$

\end{document}